\documentclass{article} 
\usepackage{iclr2027_conference,times}

\usepackage{amsmath,amsfonts,bm}

\def\eqref#1{equation~\ref{#1}}
\def\Eqref#1{Equation~\ref{#1}}

\def\1{\bm{1}}

\DeclareMathAlphabet{\mathsfit}{\encodingdefault}{\sfdefault}{m}{sl}
\SetMathAlphabet{\mathsfit}{bold}{\encodingdefault}{\sfdefault}{bx}{n}

\usepackage{microtype}
\usepackage{graphicx}
\usepackage{subcaption}
\usepackage{booktabs} 

\usepackage{hyperref}
\hypersetup{
    colorlinks=true,
    linkcolor=magenta,
    filecolor=red,
    urlcolor=cyan,
    citecolor=cyan,
}

\usepackage{xcolor}

\usepackage{amsmath}
\usepackage{amssymb}
\usepackage{mathtools}
\usepackage{amsthm}
\usepackage{latexsym}
\usepackage[T1]{fontenc}
\usepackage[utf8]{inputenc}
\usepackage{microtype}
\usepackage{graphicx}
\usepackage{url}
\usepackage{xcolor}
\usepackage{wrapfig}
\usepackage{enumitem}
\usepackage{multirow}
\usepackage{natbib}
\usepackage{booktabs}
\usepackage{arydshln}
\usepackage{appendix}
\usepackage{array}
\usepackage{stfloats}
\usepackage[table]{xcolor}
\usepackage{longtable}
\usepackage{mdframed}
\usepackage{ltablex} 
\keepXColumns
\usepackage{caption}
\usepackage{makecell} 

\usepackage{placeins}
\usepackage[normalem]{ulem}
\newcolumntype{C}[1]{>{\centering\arraybackslash}p{#1}}

\usepackage[most]{tcolorbox}
\usepackage{graphicx}

\newcounter{myboxcounter}
\renewcommand{\themyboxcounter}{\arabic{myboxcounter}}
\newtcolorbox{custombox_red}[2][]{
    colback=red!5!white,
    colframe=red!75!black,
    fonttitle=\bfseries,
    title=Box~\themyboxcounter: #2,
    enhanced,
    #1
}

\newtcolorbox{custombox_orange}[2][]{
    colback=orange!5!white,
    colframe=orange!75!black,
    fonttitle=\bfseries,
    title=Box~\themyboxcounter: #2,
    #1
}
\newtcolorbox{custombox_blue}[2][]{
    colback=blue!5!white,
    colframe=blue!75!black,
    fonttitle=\bfseries,
    title=Box~\themyboxcounter: #2,
    #1
}
\newtcolorbox{custombox_green}[2][]{
    colback=green!5!white,
    colframe=green!75!black,
    fonttitle=\bfseries,
    title=Box~\themyboxcounter: #2,
    #1
}
\newtcolorbox{custombox_black}[2][]{
    colback=black!5!white,
    colframe=black!75!black,
    fonttitle=\bfseries,
    title=Box~\themyboxcounter: #2,
    #1
}

\usepackage[most]{tcolorbox}
\newcommand{\ie}{\emph{i.e., }}
\newcommand{\eg}{\emph{e.g., }}

\newcommand{\wrt}{\emph{w.r.t. }}
\newcommand{\cf}{\emph{cf. }}

\definecolor{ForestGreen}{RGB}{34,139,34}
\renewcommand{\citet}[1]{\citeauthor{#1} [\citeyear{#1}]}

\usepackage[capitalize,noabbrev]{cleveref}

\definecolor{confrankone}{HTML}{BDDCF2}
\definecolor{confranktwo}{HTML}{DCECF8}
\definecolor{confrankthree}{HTML}{F1F7FC}
\newcommand{\crone}[1]{\cellcolor{confrankone}#1}
\newcommand{\crtwo}[1]{\cellcolor{confranktwo}#1}
\newcommand{\crthree}[1]{\cellcolor{confrankthree}#1}

\theoremstyle{plain}

\theoremstyle{definition}

\theoremstyle{remark}

\usepackage[textsize=tiny]{todonotes}

\title{\textsl{Less Data Approximates More}: Earning Faithful\\Confidence in High-Stakes Domains}

\author{
Haokai Ma$^{1}$, Lee Yan Zhen$^{1}$, Gang Yang$^{1}$, Yunxiang Chen$^{1}$,\\
\textbf{Yunshan Ma$^{2}$, Tat-Seng Chua$^{1}$, Ee-Chien Chang$^{1}$}\\
$^1$National University of Singapore,\\
$^2$Singapore Management University,\\
\texttt{haokai.ma@nus.edu.sg}, \texttt{lyanzhen@u.nus.edu}
}

\iclrfinalcopy 
\begin{document}

\maketitle

\lhead{Under review.}
\renewcommand{\headrulewidth}{0.4pt} 

\begin{abstract}
Large language models are increasingly deployed in high-stakes domains, where confident yet incorrect inferences may cause severe real-world harm, bringing the long-overlooked issue of confidence faithfulness to the forefront. A promising solution jointly optimizes unsupervised Reinforcement Learning from Internal Feedback (RLIF) with reasoning-trace-guided Reasoning Distillation (RD), yet it faces three persistent challenges, namely the scarcity of high-quality training corpora, factually unwarranted overconfidence, and erroneous updates amplified by indiscriminate fusion. Inspired by how human confidence accumulates from uncertainty to certainty, we propose Progressive Reasoning Gain (PRG) to measure whether reasoning steps progressively strengthen confidence in the final answer. Building on PRG, we introduce HyTuning, a hybrid post-training framework that adaptively reweights RD and RLIF, using scarce supervised reasoning traces as a stable anchor while exploiting abundant unlabeled queries for scalability. Experiments on several domain-specific and general benchmarks demonstrate that HyTuning improves accuracy while achieving confidence faithfulness under limited supervision, supporting a practical \textsl{``Less Data Approximates More''} effect. Our code will be released upon acceptance.
\end{abstract}

\begingroup
\let\oldaddcontentsline\addcontentsline
\renewcommand{\addcontentsline}[3]{}
\section{Introduction}
\label{sec:Introduction}
Large language models (LLMs) have been extensively deployed across real-world applications and exhibit remarkable performance~\citep{yang2024harnessing,PASsMoE,ResMerge}. Unlike general domains, high-stakes domains such as cybersecurity, finance, and medicine~\citep{chu2025domaino1s} can hardly tolerate confidently-wrong inferences (\cf Figure~\ref{fig:MotivationGraph}), which may mislead practitioners into harmful decisions and trigger severe system failures. These domains require LLMs to be accurate and express reliable confidence~\citep{bengio2025superintelligent}, where high-confidence inferences should be correct with a high likelihood while low-confidence ones can be deferred to human review. We term this property \emph{confidence faithfulness}.

When tackling high-stakes domains, existing works mainly inject domain-specific knowledge into LLMs via post-training~\citep{salahuddin2025less,ke2025demystifying} or condition their responses on information retrieved from external knowledge bases~\citep{shi2025searchragsearchengineshelpful,li2025lexrag}. These methods primarily target task accuracy while leaving LLM confidence unconstrained. Worse still, supervised post-training tend to reward guessing over acknowledging uncertainty~\citep{kalai2025language}, further hindering \emph{confidence faithfulness}. A natural remedy is LLM calibration, which aligns confidence with accuracy across the entire confidence range~\citep{geng2024survey}, yet \emph{confidence faithfulness} only requires high-confidence inferences to be reliable. Recently, Reinforcement Learning from Internal Feedback (RLIF)~\citep{zhao2025learningreasonexternalrewards} optimizes LLMs with their self-certainty as the reward, without requiring any ground-truth labels. Although RLIF boosts reasoning without labels, continued training raises its self-certainty without a matching accuracy gain, drifting toward overconfidence~\citep{zhang2025freelunchrethinkinginternal}. This naturally motivates a promising solution: \emph{hybrid post-training that combines supervised post-training and RLIF to jointly pursue accuracy and confidence faithfulness}.

\begin{figure*}[!t]
    \centering
    \includegraphics[width=0.8\linewidth]{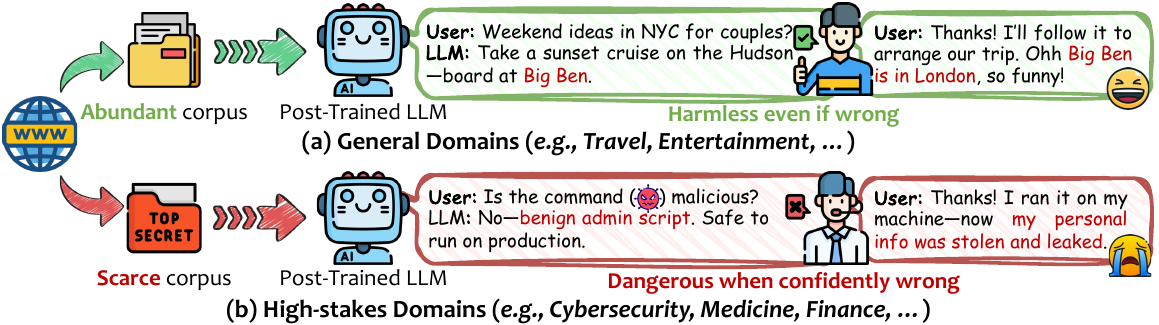}
    \vspace{-0.3cm}
    \caption{Comparison of the tolerance for confidently-wrong inferences between general domains and high-stakes domains, where the latter also suffer from scarce training corpora.}
    \vspace{-0.6cm}
    \label{fig:MotivationGraph}
\end{figure*}

Although such a hybrid solution is promising, implementing it in high-stakes domains may surface three non-trivial challenges.
1) \textbf{\emph{Scarcity of high-quality training corpora.}} In high-stakes domains, data collection is severely restricted by confidentiality and legal liability, and labeling the already scarce data (\cf Figure~\ref{fig:MotivationGraph}) relies on expensive domain experts. Even when substantial data are collected, divergent institutional practices frequently introduce noise and inconsistent annotations, further undermining their suitability for tuning reliable LLMs.
2) \textbf{\emph{Factually unwarranted overconfidence.}} Beyond pre-training that favors fluency and post-training that rewards fully-specified outputs over expressing uncertainty, high-stakes corpora are commonly framed in a definitive tone. Supervised tuning on such corpora may thus encourage assertive answers beyond what the LLM actually knows, while RLIF further sharpens such unwarranted confidence through its self-certainty reward.
3) \textbf{\emph{Indiscriminate fusion amplifies erroneous updates.}} Although hybrid post-training jointly optimizes supervised objectives and intrinsic signals, the self-certainty \wrt an LLM does not necessarily constitute a reliable optimization signal, since it can remain high along erroneous reasoning trajectories. Fusing the two objectives without discriminating the reliability of each trajectory thus reinforces confidently-wrong trajectories as readily as correct ones, which is particularly unacceptable in high-stakes domains.

To address these challenges, we propose HyTuning, a hybrid post-training framework that integrates RLIF with Reasoning Distillation (RD) to harmonize task performance and confidence faithfulness of LLMs in high-stakes domains. Specifically, RLIF derives intrinsic rewards from the self-certainty of LLMs, enabling scalable self-improvement on abundant unlabeled queries and alleviating the \emph{scarcity of high-quality training corpora}. Since the unsupervised RLIF tends to drift toward overconfidence, RD anchors the optimization to the scarce yet verified reasoning traces, grounding LLM confidence in factual correctness. We further equip RD with stochastic regularization to reduce over-specialization to these traces and the imbalance against abundant unlabeled queries, jointly mitigating \emph{factually unwarranted overconfidence}. To prevent \emph{indiscriminate fusion from amplifying erroneous updates}, we follow the assumption that ``\uline{human reasoning typically moves from uncertainty to certainty}'' and propose Progressive Reasoning Gain (PRG) to quantify whether each reasoning step progressively strengthens the LLM's confidence in the final answer, so that trajectories whose confidence does not build up along the reasoning receive low PRG. Based on PRG, we design an adaptive weight to dynamically determine when RD should dominate to suppress unreliable intrinsic signals and when RLIF should be emphasized to align high confidence with progressively supported reasoning. Theoretically, we show that under a proxy assumption on the density ratio, the PRG-induced weight yields an importance-reweighted surrogate of the entropy over self-consistent traces up to an additive constant, which justifies selective rather than unconditional posterior sharpening.

Extensive experiments on five domain-specific benchmarks across cybersecurity and finance and one general benchmark demonstrate that HyTuning outperforms all baselines under limited supervision while preserving general capabilities. Blind human evaluation by four experts further confirms that weakly grounded reasoning concentrates significantly in the low-PRG regime. Other analysis from the perspectives of surgical studies, self-driven confidence, and training dynamics verify its confidence faithfulness and ``\textsl{Less Data Approximates More}'' property. We believe that HyTuning is not a transient expedient dictated by current data limitations. Its performance keeps rising with more supervision, while its advantage is largest where supervision is scarcest.

\section{Preliminary}
\label{sec:Preliminary}
\textbf{Definition of Confidence Faithfulness.}
To address the notorious issue of confidently-wrong inference in high-stakes domains, we first define confidence faithfulness as: when an LLM expresses high confidence, its inference should have a high likelihood of being correct. This differs from standard LLM calibration in that we do not enforce low confidence to correspond to low accuracy, since an error in a high-stakes domain costs far more than human-in-the-loop and low-confidence inferences should therefore be treated with default caution. Concretely, given the query set $X\!=\!\{x_1,\cdots,x_n\}$ with ground-truth answers $Y\!=\!\{y_1,\cdots,y_n\}$, the corresponding inference set $\hat{Y}\!=\!\{\hat{y}_1,\cdots,\hat{y}_n\}$, and their associated confidence scores $C\!=\!\{c_1,\cdots,c_n\}$ derived from a certain uncertainty metric, we sort all instances in descending order of $c_i$. An LLM then exhibits \textbf{confidence faithfulness} when the accuracy over its most confident $\kappa$ fraction exceeds the accuracy over all instances and increases as $\kappa$ shrinks. Following selective prediction~\citep{geifman2018bias,traub2024overcoming}, this criterion constrains only the reliability of the high-confidence region, which is the asymmetry that real-world deployment requires. Since the ranking rather than the numerical value of $C$ determines which instances are retained, it is further invariant to any monotone rescaling of $C$, which allows different post-training strategies to be compared on the same uncertainty metric.

\textbf{Progressive Reasoning Gain within LLM Reasoning.} 
In high-stakes tasks, accuracy alone is an inadequate proxy for LLMs' reliability, since even a correct answer may rest on a reasoning trajectory with many irrelevant or self-contradictory intermediate steps. More importantly, a characteristic property of a sound reasoning trajectory is \emph{confidence accumulation}, that is, the model's confidence to generate the final answer should be progressively strengthened as the reasoning unfolds, mirroring \textbf{\emph{how human reasoning typically moves from uncertainty to certainty}}. Motivated by this, we introduce a trajectory-level metric termed Progressive Reasoning Gain (PRG) to quantify whether a reasoning trace exhibits progressive confidence accumulation (\cf Appendix~\ref{app:prg-screen}), which serves as a one-sided indicator (\cf Appendix~\ref{app:prg-screen}) that flags traces whose confidence fails to build up:
\begin{equation}
    \label{eq:prg}
   \mathcal{P} \!=\! \frac{1}{T}\!\sum\limits^{T}_{t=1}\!\mathrm{ReLU}(\log p_{\pi_{\theta}}(\hat{y} \!\mid\! q, \!o_{\leq t}) \!-\! \log p_{\pi_{\theta}}(\hat{y} \!\mid\! q, \!o_{\leq t-1})), 
\end{equation}
where $o_{1:T}$ is the generated reasoning trace, $\hat{y}$ denotes the answer that the LLM generates from $o_{1:T}$, and $\log p_{\pi_{\theta}}(\hat{y} \!\mid\! q, \!o_{\leq t})$ measures the confidence of the LLM in this answer under the first $t$ steps alone.

\section{Methodology}
\label{sec:Methodology}
We begin by casting RD and RLIF under a distributional lens (\cf Section~\ref{sec:unified_framework}). Building on this view, we introduce HyTuning, a hybrid post-training framework that combines the stability of RD with the scalability of RLIF, while explicitly controlling their respective failure modes (\cf Section~\ref{sec:HyTuning}).

\begin{wrapfigure}{R}{0.39\textwidth}
    \vspace{-0.5cm}
    \includegraphics[width=\linewidth]{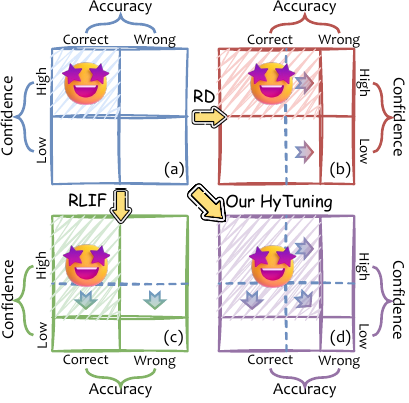}
    \vspace{-0.7cm}
    \caption{A toy example illustrating how our HyTuning goes beyond RD (accuracy-focused) and RLIF (confidence-focused) to jointly improve accuracy and confidence faithfulness.}
    \vspace{-0.6cm}
    \label{fig:ideal_performance}
\end{wrapfigure}
\subsection{Unified Framework of RD and RLIF}
\label{sec:unified_framework}
For high-stakes domains, confidence faithfulness is as important as task performance: confident but incorrect predictions may cause severe downstream harm~\citep{bengio2025superintelligent}. A key driver of poor confidence faithfulness is that standard supervised pipeline typically rewards guessing over acknowledged uncertainty, which encourages unwarranted certainty under ambiguity~\citep{kalai2025language}. Recently, label-free RL approaches that rely only on model-derived signals have been proposed to improve reasoning and self-awareness~\citep{zhao2025learningreasonexternalrewards,he2026far,EM-RL,he2026visplay}. However, continued optimization can make their output distributions increasingly confident without a corresponding gain in accuracy~\citep{zhang2025freelunchrethinkinginternal}. This raises a central question: \textit{Can we formulate a unified post-training framework that integrates RD and RLIF, turning the trade-off between task performance and confidence faithfulness into a \textbf{win--win} outcome?}

\textbf{RD \emph{vs.} RLIF for the trade-off between performance and confidence faithfulness.} RD transfers reasoning traces extracted from a teacher into the student (\cf Appendix~\ref{app:RD}). In a distributional view, it anchors the student reasoning distribution $p_{\theta}(\cdot\!\mid\! q)$ toward the teacher-induced distribution $p_{t}(\cdot\!\mid\! q)$ by:
\begin{equation}
\label{eq:rd_anchor_kl}
\min_{\theta}\; \mathbb{D}_{\mathrm{KL}}\!\left(p_t(\cdot\mid q)\,\Vert\,p_{\theta}(\cdot\mid q)\right).
\end{equation}
With high-quality supervision, RD typically improves accuracy (\cf Figure~\ref{fig:ideal_performance} (a)$\Rightarrow$(b)). Its main limitation is scalability: high-quality traces are scarce and expensive in high-stakes domains, and the accuracy of RD degrades rapidly once they are unavailable at scale (\cf Table~\ref{tab.overall_performance}).

In contrast, RLIF avoids this dependence on supervision by training on a label-free corpus with intrinsic signals derived from the LLM itself as rewards (\cf Appendix~\ref{app:RLIF}). At a high level, it can be viewed as encouraging a sharper (low-entropy) policy-induced trajectory distribution:
\begin{equation}
\label{eq:entropy_view_prior}
\min_{\theta}\; H\!\left(p_{\theta}(\cdot\mid q)\right), \quad\text{where}\quad H\!\left(p_{\theta}(\cdot\mid q)\right)= -\mathbb{E}_{o\sim p_{\theta}(\cdot\mid q)}\!\left[\log p_{\theta}(o\mid q)\right].
\end{equation}
Without ground truth, intrinsic rewards may drift away from factual correctness, leading to accuracy loss or factually unwarranted overconfidence (\cf Figure~\ref{fig:ideal_performance} (a)$\Rightarrow$(c)). These complementary strengths motivate a unified perspective that combines the grounding of RD with the scalability of RLIF rather than treating them as isolated paradigms. Yet these two objectives pull in opposite directions, as Equation~\ref{eq:rd_anchor_kl} spreads the student's distribution to cover the teacher while Equation~\ref{eq:entropy_view_prior} concentrates it.

\textbf{Design principles.} The above analysis suggests two principles for high-stakes post-training:

\textbf{Principle 1:} \emph{When fused with RD, RLIF should move beyond unconditional prior entropy minimization. Instead, it should sharpen the policy distribution only when a trajectory exhibits self-consistent confidence accumulation, denoted by a condition $z$ indicating it deserves increased confidence.}

\textbf{Principle 2:} \emph{When $z$ indicates that intrinsic feedback is likely to be unreliable, RD should provide an anchoring gradient that stabilizes learning and grounds optimization using high-quality supervision.}

These principles motivate a composite objective, where a yet-to-be-specified self-consistency condition $z$ determines whether to sharpen the policy distribution or fall back on teacher guidance:
\begin{equation}
\label{eq:composite_pseudo}
\min_{\theta}\; z\cdot H\!\left(p_{\theta}(\cdot\mid q)\right)
\;+\;
(1-z)\,\mathbb{D}_{\mathrm{KL}}\!\left(p_t(\cdot\mid q)\,\Vert\,p_{\theta}(\cdot\mid q)\right),
\end{equation}
which reduces to RLIF when $z$=1 and to RD when $z$=0. Next, we instantiate \Eqref{eq:composite_pseudo} into a practical post-training strategy that improves both task performance and confidence faithfulness.

\subsection{Proposed Method: HyTuning}
\label{sec:HyTuning}
We instantiate \Eqref{eq:composite_pseudo} as HyTuning, which learns from intrinsic feedback on abundant unlabeled queries while keeping optimization anchored to scarce reasoning traces (\cf Figure~\ref{fig:overall_structure}). For queries with verified targets, PRG decides how much the intrinsic feedback of each trace should be trusted relative to the target, so confidence grows mainly along trajectories whose reasoning supports it.
\begin{figure*}[!t]
    \centering
    \includegraphics[width=0.98\linewidth]{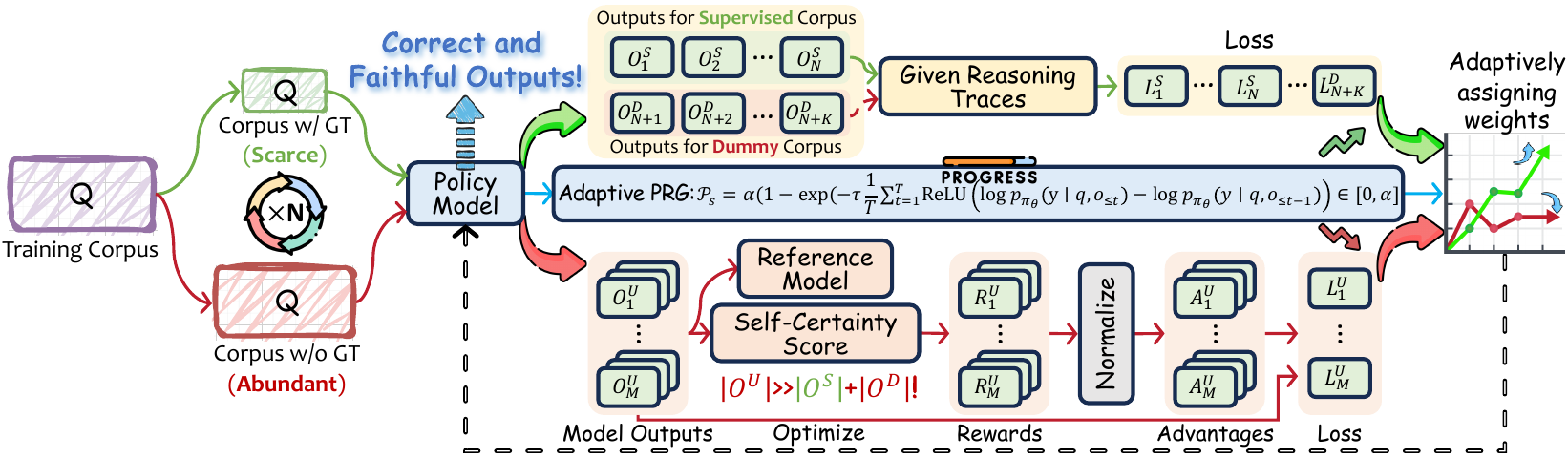}
    \vspace{-0.3cm}
    \caption{Overview of our proposed HyTuning.}
    \vspace{-0.7cm}
    \label{fig:overall_structure}
\end{figure*}

\textbf{Reasoning distillation on supervised and dummy corpora.} Each query $q$ in the supervised corpus $\mathcal{D}^{\mathrm{S}}$ is paired with a target sequence $s\!=\!(r,y)$, which concatenates a verified reasoning trace $r$ and the ground truth (GT) $y$ (\cf Appendix~\ref{app:RD}). Maximizing the likelihood of $s$ instantiates the anchoring term in \Eqref{eq:composite_pseudo}, which grounds LLM confidence in verified reasoning. However, \emph{high-quality traces are scarce in high-stakes domains}, leaving $\mathcal{D}^{\mathrm{S}}$ far smaller than the unsupervised corpus $\mathcal{D}^{\mathrm{U}}$. To mitigate this imbalance and reduce over-specialization along the supervised pathway, we introduce a dummy regularizer in the spirit of \citet{xie2016disturblabel}. Specifically, we randomly route 5\% of queries in $\mathcal{D}^{\mathrm{U}}$ to RD with pseudo target tokens, forming a dummy corpus $\mathcal{D}^{\mathrm{D}}$. The resulting noisy loss regularizes RD beyond the supervised corpus without requiring external labels. The RD objective is:
\begin{equation}
\label{eq.final_rd}
\mathcal{J}_{\mathrm{RD}}(\theta)=
\mathbb{E}_{(q,s)\sim \mathcal{D}^{\mathrm{S}}\cup\mathcal{D}^{\mathrm{D}}}\left[\frac{1}{|s|}\sum_{t=1}^{|s|}\log \pi_{\theta}\!\left(s_t \mid q, s_{<t}\right)\right].
\end{equation}
Since RLIF draws $G$ rollouts per query under GRPO, we replicate each sample in $\mathcal{D}^{\mathrm{S}}\!\cup\!\mathcal{D}^{\mathrm{D}}$ accordingly to align the update frequency of RD with that of the RLIF branch, at no additional annotation cost.

\textbf{Self-certainty optimization on the unsupervised corpus.} For each query $q$, including those in $\mathcal{D}^{\mathrm{S}}\!\cup\!\mathcal{D}^{\mathrm{D}}$, the RLIF branch samples $G$ rollouts from the old policy $\pi_{\theta_{\mathrm{old}}}$ and scores each rollout $o^{\mathrm{U}}_i$ with its self-certainty $c(o^{\mathrm{U}}_i\!\mid\! q)$~\citep{zhao2025learningreasonexternalrewards}. These rewards are normalized within the group into advantages, which drive a GRPO-style clipped surrogate $\mathcal{J}_{\mathrm{RLIF}}(\theta)$ with a KL penalty against policy drift (\cf Appendix~\ref{app:RLIF}). Since raising self-certainty sharpens the policy distribution, $\mathcal{J}_{\mathrm{RLIF}}(\theta)$ instantiates the sharpening term in \Eqref{eq:composite_pseudo} without any GT labels. While this branch scales with the abundant $\mathcal{D}^{\mathrm{U}}$, its self-certainty can remain high along erroneous trajectories, calling for a trajectory-aware mechanism that decides when intrinsic feedback should yield to RD.

\textbf{Highlighting self-consistent reasoning via an adaptive PRG score.} 
To determine how much intrinsic feedback each trajectory deserves, HyTuning emphasizes self-consistent trajectories, whose reasoning steps progressively strengthen confidence in the final answer. We quantify this property for every rollout with the PRG $\mathcal{P}$ in \Eqref{eq:prg}. \textbf{PRG is one-sided by construction.} Since confidence accumulation is a necessary rather than sufficient property of well-grounded reasoning, a near-zero $\mathcal{P}$ indicates little accumulated support for the final answer, whereas a high $\mathcal{P}$ cannot guarantee well-grounded reasoning. Our human evaluation confirms this one-sided behavior (\cf Appendix~\ref{app:prg-screen}): trajectories whose conclusions are poorly supported by their own reasoning concentrate in the low-PRG regime, whereas higher $\mathcal{P}$ does not consistently indicate better-grounded reasoning. Accordingly, rather than feeding the unbounded $\mathcal{P}\!\in\![0,\!+\infty)$ into the objective, we map it to a bounded and saturating weight $\mathcal{P}_{\mathrm{s}}\!\in\![0,\alpha]$ that rises quickly for small $\mathcal{P}$ and levels off toward $\alpha$ as $\mathcal{P}$ grows:
\begin{equation}
\label{eq.weight}
\mathcal{P}_{\mathrm{s}} = \alpha \cdot \left(1 - \exp(-\tau \cdot \mathcal{P})\right),
\end{equation}
where $\alpha\!\in\![0,1]$ caps the maximum influence of intrinsic feedback, and the temperature $\tau$ controls how quickly $\mathcal{P}_{\mathrm{s}}$ approaches this cap. This mapping preserves discriminability in the low-PRG regime where PRG is informative, and assigns nearly uniform weights in the high-PRG regime where it is not. We then connect $\mathcal{P}_{\mathrm{s}}$ to the condition $z$ in \Eqref{eq:composite_pseudo}. Since no explicit binary self-consistency label is observed, we treat $z$ as a latent event and use $\mathcal{P}_{\mathrm{s}}$ (an unnormalized evidence score for $z$) as a bounded monotone proxy for the density ratio between the posterior and the prior in Bayes' rule, \ie
\begin{equation}
\label{eq:ps_ratio_proxy}
\frac{p_\theta(o\mid q,z)}{p_\theta(o\mid q)}=\frac{p_\theta(z\mid q,o)}{p_\theta(z\mid q)}\approx k\,\mathcal{P}_{\mathrm{s}},
\end{equation}
where the proportional constant $k$ only introduces a global scaling, so that $\mathcal{P}_{\mathrm{s}}$ itself acts as an importance weight that allows expectations under the posterior to be estimated from the rollouts sampled from the prior (derivation in Appendix~\ref{app-sec:proof}). Under this view, RLIF shifts from sharpening the prior $p_{\theta}(\cdot\!\mid\! q)$ in \Eqref{eq:entropy_view_prior} to selectively sharpening the $z$-conditioned posterior $p_{\theta}(\cdot\!\mid\! q,z)$. In particular, the posterior entropy can be approximated by the following importance-weighted form:
\begin{equation}
\label{eq:posterior_entropy_approx}
\max_{\theta}\;H\!\left(p_{\theta}(\cdot\mid q,z)\right)\propto-\mathbb{E}_{o\sim p_{\theta}(\cdot\mid q)}\!\left[\mathcal{P}_{\mathrm{s}}\,\log p_{\theta}(o\mid q)\right]+ \mathrm{const.},
\end{equation}
which differs from the prior entropy in \Eqref{eq:entropy_view_prior} only by the trajectory-dependent weight $\mathcal{P}_{\mathrm{s}}$ inside the expectation. Consequently, $\mathcal{P}_{\mathrm{s}}$ realizes the selective sharpening of Principle~1 by suppressing it along weakly grounded trajectories while retaining it along those whose confidence accumulates progressively. The intrinsic feedback withdrawn in this way is compensated by a stronger RD anchor as prescribed by Principle~2, which we instantiate in the hybrid objective below.

\textbf{Hybrid learning objective.}
Equipped with $\mathcal{P}_{\mathrm{s}}$, for each query with a stored target in $\mathcal{D}^{\mathrm{S}}\!\cup\!\mathcal{D}^{\mathrm{D}}$, we combine the RLIF objective on each of its rollouts with the RD objective on the target:
\begin{equation}
\label{eq:final_obj}
\mathcal{L}=\mathcal{P}_{\mathrm{s}}\,\mathcal{J}_{\mathrm{RLIF}}(\theta)+(1-\mathcal{P}_{\mathrm{s}})\,\mathcal{J}_{\mathrm{RD}}(\theta),
\end{equation}
where $\mathcal{P}_{\mathrm{s}}$ is computed from the rollout, and $\mathcal{J}_{\mathrm{RLIF}}(\theta)$ and $\mathcal{J}_{\mathrm{RD}}(\theta)$ denote the per-sample terms of \Eqref{eq.rlif} and \Eqref{eq.final_rd} on the rollout and its paired target, respectively. The overall objective averages $\mathcal{L}$ over all rollouts in a batch. Queries in $\mathcal{D}^{\mathrm{U}}$ have no stored target and are optimized with $\mathcal{J}_{\mathrm{RLIF}}(\theta)$ alone. Intrinsic feedback is thus concentrated on trajectories with accumulated confidence, while weakly-grounded trajectories defer to the verified target (\cf Figure~\ref{fig:ideal_performance} (a)$\Rightarrow$(d)).

\section{Experiments}
\label{sec:Experiments}
\subsection{Experimental Setup}

\textbf{Evaluation protocol.} To evaluate the effectiveness of HyTuning, we utilize four benchmarks from both the cybersecurity domain (\ie CSEBenchmark~\citep{wang2025digitalcybersecurityexpertfar}, CyberMetric-500~\citep{tihanyi2024cybermetricbenchmarkdatasetbased}, and ASBench~\citep{ma2025attackseqbenchbenchmarkinglargelanguage}) and the general domain (\ie MMLU~\citep{hendrycks2021measuringmassivemultitasklanguage}), and further extend the evaluation to two finance benchmarks in Section~\ref{sec.universality} to show that HyTuning is not confined to a single high-stakes domain. Further details are listed in Appendix~\ref{appendix:evaluation_protocol}.

\textbf{Baselines.} We compare HyTuning against eight post-training baselines (\ie Supervised Fine-Tuning (SFT)~\citep{ouyang2022traininglanguagemodelsfollow}, RD~\citep{RD}, Group Relative Policy Optimization (GRPO)~\citep{shao2024deepseekmathpushinglimitsmathematical}, INTUITOR~\citep{zhao2025learningreasonexternalrewards}, Reinforcement Learning with Reference Probability Reward (RLPR)~\citep{yu2025rlprextrapolatingrlvrgeneral}, Hybrid Post-Training (HPT)~\citep{lv2025unifiedviewlargelanguage}, On-Policy Distillation (OPD)~\citep{OPD}, and On-Policy Self-Distillation (OPSD)~\citep{OPSD}) to investigate its effectiveness and confidence faithfulness (\cf Appendix~\ref{appendix:baselines}). 

\textbf{Evaluation Metrics.} We jointly evaluate HyTuning and the post-training baselines along two dimensions, namely task performance and model confidence, whose consistency constitutes the confidence faithfulness formalized in Section~\ref{sec:Preliminary}. Further details can be found in Appendix~\ref{appendix:metrics}.

\begin{table*}[!t]
\centering
\caption{Performance comparison on domain-specific and general benchmarks. }
\vspace{-0.3cm}
\setlength{\tabcolsep}{4pt}
\resizebox{\linewidth}{!}{
\label{tab.overall_performance}
\begin{tabular}{c|ccccc|ccccc}
\toprule
\multirow{3}{*}{\textbf{Models}}&\multicolumn{5}{c|}{\textbf{Training w/ Partial Supervision}} & \multicolumn{5}{c}{\textbf{Training w/ Full Supervision}} \\\cmidrule(lr){2-6}\cmidrule(lr){7-11}
& \multicolumn{3}{c}{\textbf{Domain-specific Bench.}} & \textbf{Gen. Bench.} & \multirow{2}{*}{\textbf{Aver.}} & \multicolumn{3}{c}{\textbf{Domain-specific Bench.}} & \textbf{Gen. Bench.} & \multirow{2}{*}{\textbf{Aver.}} \\ \cmidrule(lr){2-4}\cmidrule(lr){5-5}\cmidrule(lr){7-9}\cmidrule(lr){10-10}
& \textbf{ASB} &\textbf{CSEB} & \textbf{CyberM} & \textbf{MMLU} &  & \textbf{ASB} &\textbf{CSEB} & \textbf{CyberM} & \textbf{MMLU} & \\
\midrule 
Qwen2.5-3B & 0.269 & 0.318 & 0.556 & 0.449 & 0.398 & 0.269 & 0.318 & 0.556 & 0.449 & 0.398 \\ \midrule
SFT~\citep{ouyang2022traininglanguagemodelsfollow} & 0.102 & 0.133 & 0.260 & 0.256 & 0.188 & 0.305 & 0.286 & 0.554 & 0.447 & 0.398  \\
RD~\citep{RD} & 0.460 & 0.328 & 0.666 & 0.457 & 0.478 & \underline{0.516} & 0.435 & 0.748 & 0.529 & 0.557  \\
GRPO~\citep{shao2024deepseekmathpushinglimitsmathematical} & 0.437 & 0.348 & 0.660 & \underline{0.521} & 0.492 & 0.499 & \underline{0.449} & \underline{0.770} & \textbf{0.562} & \textbf{0.570}  \\
INTUITOR~\citep{zhao2025learningreasonexternalrewards} & \underline{0.476} & 0.321 & 0.700 & 0.476 & \underline{0.494} & 0.475 & 0.334 & 0.752 & 0.465 & 0.507  \\
RLPR~\citep{yu2025rlprextrapolatingrlvrgeneral} & 0.414 & \underline{0.353} & \underline{0.706} & 0.475 & 0.487 & 0.488 & \textbf{0.451} & \textbf{0.774} & 0.483 & 0.549  \\
HPT~\citep{lv2025unifiedviewlargelanguage} & 0.253 & 0.224 & 0.408 & 0.413 & 0.325 & 0.463 & 0.408 & 0.740 & 0.498 & 0.527  \\
OPD~\citep{OPD} & 0.271 & 0.269 & 0.520 & 0.419 & 0.370 & 0.327 & 0.357 & 0.596 & 0.493 & 0.443  \\
OPSD~\citep{OPSD} & 0.322 & 0.291 & 0.554 & 0.439 & 0.401 & 0.385 & 0.214 & 0.432 & 0.398 & 0.357  \\ \midrule
\cellcolor{yellow!10} \textbf{HyTuning (Ours)} & \cellcolor{yellow!10} \textbf{0.489} & \cellcolor{yellow!10} \textbf{0.391}& \cellcolor{yellow!10} \textbf{0.732} & \cellcolor{yellow!10} \textbf{0.552} & \cellcolor{yellow!10} \textbf{0.541} & \cellcolor{yellow!10} \textbf{0.526} & \cellcolor{yellow!10} 0.437 & \cellcolor{yellow!10} 0.766 & \cellcolor{yellow!10} \underline{0.540} & \cellcolor{yellow!10} \underline{0.567}  \\ \bottomrule
\end{tabular}}
\vspace{-0.6cm}
\end{table*}
\subsection{Effectiveness of HyTuning}
\label{sec.PerformanceComparison}
Table~\ref{tab.overall_performance} reports the comparison under two supervision regimes. Under partial supervision, each baseline accesses only the 100 supervised instances or the 3,000 synthetic queries according to its data requirement (cf. Appendix~\ref{appendix:training_corpus}), whereas HyTuning consumes both and attains the best accuracy on all benchmarks. It improves the second-best method by 9.5\% relatively in average accuracy and exceeds both constituents RD (0.478) and INTUITOR (0.494). Under full supervision, HyTuning remains competitive, attaining the best accuracy on ASBench and trailing GRPO by only 0.003 in average accuracy while still surpassing both constituents. Besides, the comparison across the two regimes further exposes a ``\textsl{Less Data Approximates More}'' property. When the supervised corpus shrinks to roughly \textbf{3\%}, the strongest fully supervised baselines GRPO, RD, and RLPR lose 0.062 to 0.079 in average accuracy, whereas HyTuning loses only 0.026. This implies that \textbf{HyTuning recovers most of the benefit of abundant supervision from only a small fraction of it, which is critical in high-stakes domains where annotation cost is the binding constraint}. Results on additional backbones and other high-stakes domains are reported in Section~\ref{subsec.scaling_llm} and Section~\ref{sec.universality}.

\subsection{Confidence Faithfulness of HyTuning}
\label{sec:conf_faith}
Beyond effectiveness, confidence faithfulness is equally critical in high-stakes domains. To examine it for HyTuning and the baselines, we represent confidence with Shifting Attention to Relevance (SAR)~\citep{duan2024shifting} and Self-Certainty (SC)~\citep{zhao2025learningreasonexternalrewards} (\cf Appendix~\ref{appendix:confidence_estimation}), and report hAURC@$\kappa$ within the most confident $\kappa\!\in\!$\{5\%,10\%,20\%\} fraction of CyberMetric in Table~\ref{tab:faithfulness} (\cf Appendix~\ref{appendix:metrics}). Since SC is the intrinsic reward of both INTUITOR and the RLIF branch of HyTuning, we regard SAR as the more objective measure and SC as a complementary reference for whether the optimized signal remains faithful. Under SAR, HyTuning and INTUITOR exhibit opposite trends as $\kappa$ shrinks from 20\% to 5\%, where the hAURC of the former drops from 17.20 to 
\begin{wraptable}{r}{0.55\textwidth}
\centering
\footnotesize
\caption{Confidence faithfulness (hAURC$\downarrow$) under SAR and SC at $\kappa\!\in\!$\{5\%,10\%,20\%\}. Shading ranks the three regions per method and measure: \colorbox{confrankone}{1st}, \colorbox{confranktwo}{2nd}, and \colorbox{confrankthree}{3rd}.}
\vspace{-0.3cm}
\label{tab:faithfulness}
\setlength{\tabcolsep}{4pt}
\resizebox{\linewidth}{!}{
\begin{tabular}{l ccc ccc}
\toprule
\multirow{2}{*}{\textbf{Models}} & \multicolumn{3}{c}{\textbf{SAR}} & \multicolumn{3}{c}{\textbf{SC}} \\
\cmidrule(lr){2-4} \cmidrule(lr){5-7}
 & 5\% & 10\% & 20\% & 5\% & 10\% & 20\% \\
\midrule
Backbone & \crthree{65.04} & \crtwo{60.41} & \crone{51.20} & \crone{45.30} & \crthree{47.04} & \crtwo{45.94} \\
RD & \crone{23.76} & \crtwo{28.30} & \crthree{31.32} & \crone{28.62} & \crtwo{31.98} & \crthree{33.83} \\
INTUITOR & \crthree{28.30} & \crtwo{25.42} & \crone{24.87} & \crthree{24.94} & \crtwo{19.63} & \crone{18.22} \\
GRPO & \crone{36.67} & \crthree{39.99} & \crtwo{38.27} & \crone{20.35} & \crtwo{30.72} & \crthree{34.47} \\
OPSD & \crthree{74.69} & \crtwo{70.19} & \crone{68.89} & \crthree{59.70} & \crtwo{58.27} & \crone{57.37} \\
HyTuning & \crone{\textbf{6.77}} & \crtwo{\textbf{11.80}} & \crthree{\textbf{17.20}} & \crtwo{\textbf{12.40}} & \crone{\textbf{12.17}} & \crthree{\textbf{12.53}} \\
\bottomrule
\end{tabular}}
\vspace{-0.5cm}
\end{wraptable}
6.77 while that of the latter rises from 24.87 to 28.30. This \emph{high-confidence risk} of INTUITOR is consistent with the overconfidence tendency reported by~\citet{zhang2025freelunchrethinkinginternal}, whereas HyTuning progressively excludes errors from its most confident inferences and attains the lowest hAURC in every region under both measures. These results indicate that \textbf{HyTuning avoids the high-confidence risk of pure RLIF, keeping its most confident inferences increasingly reliable under SAR as the region narrows}. Complete results with other metrics are provided in Appendix~\ref{appendix:faithfulness_region}.

\begin{wrapfigure}{R}{0.3\textwidth}
    \vspace{-0.5cm}
    \centering
    \includegraphics[width=\linewidth]{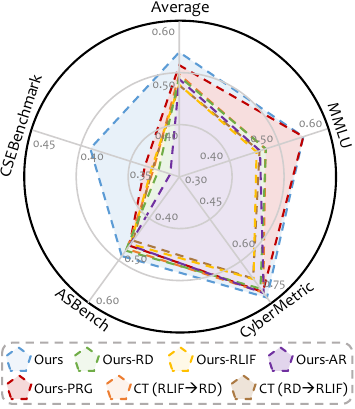}
    \vspace{-0.6cm}
    \caption{Ablation results of HyTuning, where each axis is scaled independently.}
    \vspace{-0.5cm}
    \label{fig:Ablation_Study}
\end{wrapfigure}
\subsection{Surgical Analysis of HyTuning}
\label{sec:surgical_analysis}
To investigate the contribution of each component within HyTuning, we implement multiple ablation variants in Figure~\ref{fig:Ablation_Study}. Here, ``Ours-RD'' and ``Ours-RLIF'' remove the corresponding constituent strategy, ``Ours-AR'' removes the adaptive reweighting and combines RD and RLIF with equal weight, ``Ours-PRG'' replaces PRG with self-certainty as the adaptive signal, and ``CT'' applies the two constituent strategies sequentially. Removing either constituent strategy or the adaptive reweighting mechanism degrades accuracy on every benchmark, which confirms that both the hybridization and its weighting are indispensable. Notably, ``Ours-AR'' even falls below ``Ours-RD'' in average accuracy, indicating that fusing the two strategies with a static weight degrades the stronger constituent rather than combining their strengths. Besides, ``Ours-PRG'' is the strongest ablation variant on average and matches HyTuning on MMLU, yet it underperforms HyTuning on every domain-specific benchmark. Since the two variants differ only in the signal that drives the adaptive weight, this gap isolates the contribution of PRG. Meanwhile, each CT variant stays close to the strategy applied in its first stage and falls well below HyTuning, showing that sequential tuning fails to integrate the two strategies. Complete results and further analysis are listed in Appendix~\ref{appendix:com_results}.

\begin{figure*}[!t]
    \centering
    \includegraphics[width=0.98\linewidth]{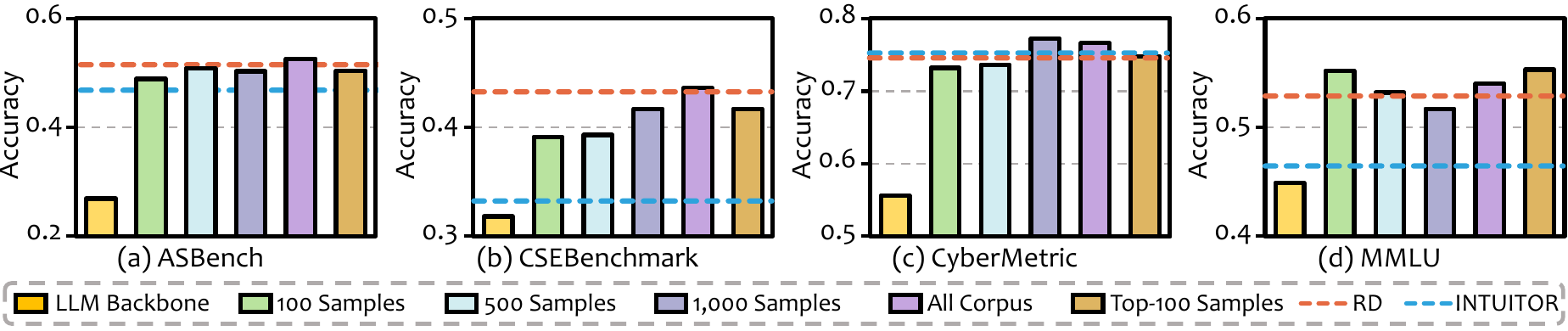}
    \vspace{-0.3cm}
    \caption{Impact of supervised data volume and quality on HyTuning across four benchmarks, where the dashed lines denote RD and INTUITOR trained on the complete corpus.}
    \vspace{-0.7cm}
    \label{fig:PotentialScaling}
\end{figure*}

\subsection{Scaling Potential of HyTuning}
\label{sec.scaling_potential}
\textbf{Scaling Potential \wrt Supervised Data Volume and Quality.}
We analyze how the volume and quality of supervised data affect HyTuning in Figure~\ref{fig:PotentialScaling}, where the two dashed lines report RD and INTUITOR trained on the complete corpus. Enlarging the supervised data from 100 to all-available samples lifts ASBench from 0.489 to 0.526, CSEBenchmark from 0.391 to 0.437 and CyberMetric from 0.732 to 0.766, confirming the \textsl{scaling potential} of HyTuning on domain-specific tasks. Under the complete corpus, HyTuning further exceeds RD and INTUITOR on every benchmark, indicating that \textbf{the benefit of HyTuning scales with supervision rather than being confined to the low-supervision regime}. Every configuration of HyTuning remains above its backbone and INTUITOR on MMLU, and the 100-sample setting even attains the peak results beyond fully supervised RD. It is reasonable since the RD branch of HyTuning simultaneously anchors the RLIF branch and injects domain-specific knowledge. The anchoring role is already fulfilled with around 100 samples, so additional traces mainly inject more domain-specific knowledge, which improves domain accuracy at the cost of general capability. Besides, selecting the 100 traces by quality rather than at random improves all domain-specific benchmarks, and matches the 1,000-sample setting on ASBench and CSEBenchmark with one tenth of the supervision (\cf Appendix~\ref{app:top100_training_corpus}). Notably, \textbf{the 100 randomly-selected samples already recover 90\% to 98\% of the accuracy of fully supervised RD on domain-specific benchmarks}, which is precisely the ``\textsl{Less Data Approximates More}'' property.

\label{subsec.scaling_llm}
\begin{wraptable}{r}{0.4\textwidth}
\vspace{-0.45cm}
\caption{Performance comparison on three domain-specific benchmarks. }
\centering
\footnotesize
\vspace{-0.25cm}
\setlength{\tabcolsep}{2.4pt}
\label{tab.scaling_llm}
\resizebox{\linewidth}{!}{
\begin{tabular}{c|cccc}
\toprule
\textbf{Models} & \textbf{ASB} &\textbf{CSEB} & \textbf{CyberM} & \textbf{Aver.} \\
\midrule 
Qwen2.5-7B & 0.495 & 0.460 & 0.708 & 0.554 \\
\quad+ HyTuning & 0.514 & 0.480 & 0.788 & 0.594 \\ \midrule
Qwen2.5-14B & 0.541 & 0.551 & 0.818 & 0.637 \\
\quad+ HyTuning & 0.555 & 0.602  & 0.868 & 0.675 \\ \midrule
LLaMA3.2-3B & 0.249 & 0.195 & 0.636 & 0.360 \\
\quad+ HyTuning & 0.372 & 0.278 & 0.650 & 0.433 \\ \midrule
Ministral3-3B & 0.515 & 0.382 & 0.774 & 0.557 \\
\quad+ HyTuning & 0.521 & 0.457 & 0.780 & 0.586 \\ \bottomrule
\end{tabular}}
\vspace{-0.4cm}
\end{wraptable}
\textbf{Scaling Potential \wrt LLM Backbone.} To verify that HyTuning generalizes beyond Qwen2.5-3B, we apply it to two larger Qwen2.5 models and two 3B models from different LLM families in Table~\ref{tab.scaling_llm}. HyTuning improves every backbone on every domain-specific benchmark. Within the Qwen2.5 family, the average performance of the backbone itself rises with model size as 0.381 $\!\rightarrow\!$ 0.554 $\!\rightarrow\!$ 0.637 for 3B, 7B, and 14B, and HyTuning lifts all the three while preserving the same monotone trend as 0.537 $\!\rightarrow\!$ 0.594 $\!\rightarrow\!$ 0.675. The gains also transfer to LLaMA3.2-3B and Ministral3-3B, which indicates that HyTuning is not tied to a particular pre-training recipe. By recovering roughly 90\% of the gap between the Qwen2.5-3B and Qwen2.5-7B backbones without enlarging the model, HyTuning not only further verifies its ``\textsl{Less Data Approximates More}'' property but also offers an economical substitute for scaling when domain data is scarce. Coupled with the data-volume trends, these results position HyTuning as a paradigm for the supervision-constrained regime rather than a stopgap for small models. Its absolute performance continues to rise with backbone size, whereas its margin over existing strategies is widest exactly where labelled reasoning traces are hardest to obtain.

\begin{wraptable}{r}{0.44\textwidth}
\vspace{-0.4cm}
\centering
\footnotesize
\caption{Performance comparison on financial and general benchmarks.}
\vspace{-0.3cm}
\setlength{\tabcolsep}{3pt}
\label{tab.finance}
\resizebox{\linewidth}{!}{
\begin{tabular}{C{1.75cm}|*{2}{C{0.92cm}}|C{1.05cm}|C{0.92cm}}
\toprule
\textbf{Models} & FinQA & FinRea & MMLU & \textbf{Average} \\ \midrule
Qwen2.5-3B & 0.195 & 0.092 & 0.449 & 0.245 \\\midrule
RD & 0.269 & 0.076 & 0.494 & 0.281 \\
INTUITOR & \underline{0.385} & \underline{0.302} & \underline{0.517} & \underline{0.401} \\
HyTuning & \textbf{0.471} & \textbf{0.383} & \textbf{0.561} & \textbf{0.472} \\ \bottomrule
\end{tabular}}
\vspace{-0.4cm}
\end{wraptable}
\subsection{Universality of HyTuning}
\label{sec.universality}
All domain-specific evaluation are confined to cybersecurity, whereas confidently wrong inferences also matter in other high-stakes domains. To verify whether HyTuning's benefits arise from the coupling between RD and RLIF rather than cybersecurity-specific corpora, we re-instantiate the pipeline on the financial domain with 100 financial reasoning traces and 3,000 synthetic financial queries and evaluate on FinQA~\citep{Finqa} and FinanceReasoning~\citep{Financereasoning} together with MMLU. As reported in Table~\ref{tab.finance}, RD only raises the average accuracy on the two financial benchmarks from 0.144 to 0.173 and even degrades on FinanceReasoning, indicating that 100 supervised traces are far from sufficient to induce financial reasoning. In contrast, INTUITOR capitalizes on the abundant unlabeled queries and reaches 0.344, while HyTuning further lifts it to 0.427. This indicates that the anchoring branch contributes by grounding intrinsic updates rather than directly improving task accuracy, which is consistent with Principle~2. In addition, HyTuning retains the highest MMLU accuracy under our unified evaluation protocol, showing that this financial adaptation does not erode general capability. These observations demonstrate that \textbf{the effectiveness of HyTuning is not confined to cybersecurity but extends to high-stakes domains with distinct reasoning characteristics}.

\subsection{Training Dynamics Analysis}
\label{sec:training_dynamic}
To verify that HyTuning consistently mitigates RLIF-induced overconfidence and improves task performance under limited supervision, we track the accuracy and self-certainty of RD, RLIF, and HyTuning in Figure~\ref{fig:perf_conf_dynamics} and report the corresponding PRG and transitional word dynamics in Appendix~\ref{appendix:training_dynamic}.

\begin{wrapfigure}{R}{0.53\textwidth}
    \vspace{-0.45cm}
    \centering
    \includegraphics[width=\linewidth]{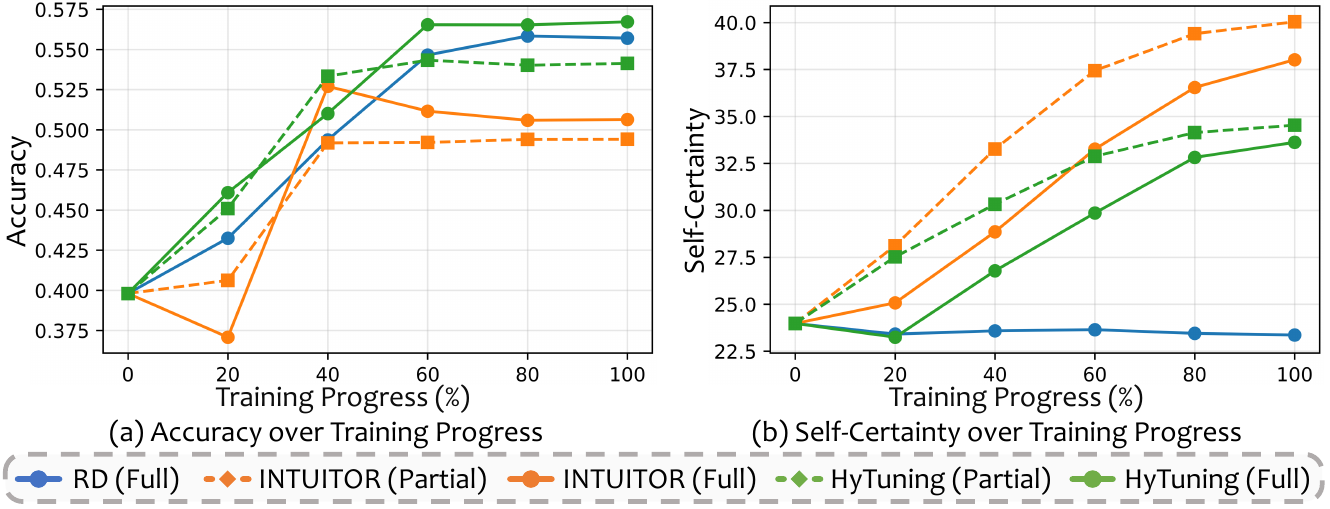}
    \vspace{-0.7cm}
    \caption{Comparison of accuracy and self-certainty over training, where Partial and Full denote the 100-sample and the complete supervised corpus.}
    \vspace{-0.3cm}
    \label{fig:perf_conf_dynamics}
\end{wrapfigure}
\textbf{Evolution of Accuracy over Training.} As shown in Figure~\ref{fig:perf_conf_dynamics}~(a), INTUITOR trained on the complete corpus suffers a pronounced early-stage drop before recovering, whereas its partial-corpus version rises monotonically yet plateaus at the lowest final accuracy among all compared settings. In contrast, both HyTuning variants and RD improve steadily throughout training. Notably, HyTuning under partial supervision approaches RD trained on the complete corpus and consistently outperforms INTUITOR (Partial), indicating that it stabilizes the optimization process and mitigates the performance degradation induced by intrinsic reward, thereby making ``\textsl{Less Data Approximates More}'' possible.

\textbf{Evolution of Self-certainty over Training.} The self-certainty of INTUITOR rises steeply on both the partial and complete supervised corpora in Figure~\ref{fig:perf_conf_dynamics}~(b), whereas that of RD stays flat and appears decoupled from optimization, which limits the reliability of its confidence faithfulness. In contrast, the self-certainty of HyTuning also rises, yet it stays below that of INTUITOR at every stage and the margin widens as training proceeds. The contrast is sharpest under partial supervision, where INTUITOR ends with the highest self-certainty and the lowest accuracy among all compared settings.

\section{Related Works}
\label{sec:Related_Works}

\noindent\textbf{Post-Training Strategies.}
Post-training is central for improving the domain-specific intelligence of LLMs beyond pre-training~\citep{zhang2024rest,zhang2025tdrm,yang2026revisiting}. Dominant paradigms such as SFT~\citep{ouyang2022traininglanguagemodelsfollow}, RD~\citep{yu2025primuspioneeringcollectionopensource, shridhar-etal-2023-distilling}, and RLVR~\citep{shao2024deepseekmathpushinglimitsmathematical, lv2025unifiedviewlargelanguage} rely on curated supervision with GT answers or reasoning traces and thus scale poorly due to annotation cost. RLIF~\citep{zhao2025learningreasonexternalrewards, zhang2025freelunchrethinkinginternal} enables label-free self-improvement via intrinsic rewards, but may suffer from  fragile late-stage dynamics. Recent works also explore combining SFT and RL under a unified objective to fully unlock the potential of LLMs~\citep{lv2025unifiedviewlargelanguage}, which are primarily tailored for mathematical reasoning with abundant supervision, limiting their applicability to high-stakes tasks where high-quality corpus are scarce. In contrast, HyTuning targets high-stakes settings where high-quality reasoning traces are inherently scarce: it couples RD and RLIF through a PRG-guided adaptive weighting mechanism, using RD to stabilize updates when intrinsic feedback is unreliable while retaining RLIF scalability.

\noindent\textbf{Confidence Estimation of LLMs.}
Confidence estimation underpins the \emph{confidence faithfulness} of LLMs, which we define as the reliability of their high-confidence inferences (\cf Section~\ref{sec:preliminary}). Confidence faithfulness is as critical as accuracy in high-stakes tasks, since systematic overconfidence can cause severe downstream harm~\citep{bengio2025superintelligent}. Prior work estimates confidence via token-level entropy~\citep{farquhar2023semanticentropy}, calibration metrics~\citep{minderer2021revisitingcalibrationmodernneural}, and probabilistic approximations~\citep{xiong2024llmsexpressuncertaintyempirical}, yet LLMs often remain miscalibrated and overconfident on incorrect answers~\citep{kadavath2022languagemodelsmostlyknow, jiang2021knowlanguagemodelsknow}. INTUITOR directly uses self-estimated confidence as a reward signal for optimization~\citep{zhao2025learningreasonexternalrewards}, which may amplify miscalibration when applied in isolation. Meanwhile, RD can correct error reasoning over standard SFT~\citep{wei2023chainofthoughtpromptingelicitsreasoning, lightman2023letsverifystepstep}, motivating our hybrid view: \emph{RD stabilizes internal rewards, while RLIF preserves adaptive self-monitoring.}

\noindent\textbf{LLMs for High-Stakes Domains.}
The growing deployment of LLMs in high-stakes domains has spurred strategies such as \emph{domain-specific tuning} and \emph{retrieval-augmented inference} to improve accuracy. \emph{Domain-specific tuning} specializes LLMs via post-training by aligning LLM priors and task formulations with expert data, yielding substantial gains in healthcare, law, and finance~\citep{singhal2022largelanguagemodelsencode, singhal2023expertlevelmedicalquestionanswering, guha2023legalbenchcollaborativelybuiltbenchmark, wu2023bloomberggptlargelanguagemodel}, but it hinges on scarce high-quality corpora. \emph{Retrieval-augmented inference} leverages Retrieval-augmented generation (RAG) to augment parametric knowledge with external evidence, improving factuality in practice. A recent work further investigate how retrieval affects LLM's confidence and improve its accuracy and expressed certainty in the medical domain~\citep{ozaki2025understandingimpactconfidenceretrieval}. Although it can mitigate certain hallucination issues, its reliance on high-quality retrieval indices prevents it from guaranteeing well-calibrated confidence.
\section{Conclusion}
\label{sec:Conclusion}
This work investigated confidence faithfulness in high-stakes domains, which requires high-confidence inferences of LLMs to be correct with a high likelihood. Inspired by how human reasoning moves from uncertainty to certainty, we proposed PRG to measure whether reasoning steps progressively strengthen confidence in the final answer. Human evaluation confirmed that weakly grounded reasoning concentrates significantly in the low-PRG regime, positioning PRG as a one-sided reliability indicator. Based on PRG, we proposed HyTuning, a hybrid post-training framework that adaptively weights RLIF against RD to harmonize performance with confidence faithfulness. Theoretically, under a proxy assumption on the density ratio, this weight yields an importance-weighted surrogate of the posterior entropy, justifying selective rather than unconditional sharpening. Extensive experiments showed that HyTuning improves accuracy and confidence faithfulness under limited supervision, supporting a practical ``\textsl{Less Data Approximates More}'' property.

\endgroup

\clearpage
\section*{AI Use Statement}
In this work, we utilize LLMs to polish the manuscript for readability and concision. For implementation, we build HyTuning upon the verl-based codebase of INTUITOR and carry out most of HyTuning and the baseline adaptations ourselves. We utilize LLMs to assist with parts of the code and manually verify all the AI-assisted code. Beyond this assistance, LLMs are also inherent experimental components since this work investigates the post-training strategies of LLMs. They serve as the post-trained backbones, the teachers of OPD and OPSD, and the reasoning models that generated the public reasoning traces in Primus-Reasoning and FinCoT. Moreover, we utilize Qwen2.5-32B-Instruct to generate the synthetic unlabeled queries for partial supervision setting (\cf Appendix~\ref{appendix:training_corpus}) and GPT-5 to serve as the judge that applies the expert-specified criteria to select the Top-100 supervised traces (\cf Appendix~\ref{app:top100_training_corpus}). We have reviewed all AI-assisted content, cross-checked the numbers stated in the text against the tables and training logs, and take full responsibility for the content of this work.

\section*{Ethics Statement}
We have read and adhere to the ICLR Code of Ethics. This work aims to reduce the confidently-wrong inferences of LLMs in high-stakes domains, where such inferences may mislead practitioners into harmful decisions. All training corpora are built upon publicly available datasets (i.e., Primus-Reasoning and FinCoT), and all evaluations are conducted on publicly available benchmarks. The cybersecurity corpora target analysis tasks such as CWE mapping, CVSS scoring, MITRE ATT\&CK mapping, and CTI understanding rather than the generation of attack payloads. The synthetic queries are generated by an LLM from these public seeds and contain no personal information. The human evaluation in Appendix~\ref{app:prg-screen} involves four experts who rate model-generated reasoning traces, and the Top-100 selection in Appendix~\ref{app:top100_training_corpus} involves three cybersecurity experts who specify the judging criteria. Both studies collect no personal information beyond the expert ratings, and all experts participate voluntarily. Finally, we note that HyTuning improves rather than guarantees confidence faithfulness. Under SAR, 4 of its 25 most confident inferences on CyberMetric-500 remain incorrect (\cf Appendix~\ref{appendix:faithfulness_region}), and PRG only flags weakly grounded trajectories rather than certifying sound ones (\cf Appendix~\ref{app:prg-screen}). Therefore, HyTuning should complement rather than replace human review in high-stakes deployments, where low-confidence inferences should still be deferred to practitioners.

\section*{Reproducibility Statement}
Our code and datasets will be available upon acceptance. Within the paper, Appendix~\ref{app-sec:proof} provides the complete derivation of the PRG-based importance reweighting, and Appendix~\ref{app:Constituent} formalizes RD and RLIF. Appendices~\ref{appendix:baselines} and~\ref{appendix:ID_baselines} detail the baselines and their training configurations, where six baselines share the verl codebase of HyTuning while RLPR and HPT follow their official codebases. Appendix~\ref{appendix:ID_HyTuning} lists all hyperparameters of HyTuning, and Appendix~\ref{appendix:Computational_Resource} reports the computational resources. Appendices~\ref{appendix:training_corpus} and~\ref{app:top100_training_corpus} specify the filtering pipeline, the supervision regimes, and the Top-100 selection, with the corresponding prompt templates in Appendix~\ref{app-sec:prompt}. Appendices~\ref{appendix:evaluation_protocol}, \ref{appendix:confidence_estimation}, and~\ref{appendix:metrics} specify the evaluation protocol, the confidence estimation, and the metrics, where accuracy is evaluated with vLLM under an identical prompt template and greedy decoding. Appendix~\ref{app:prg-screen} describes the rating criteria, the inter-expert agreement, and the statistical tests of the human evaluation. Since the rollouts are sampled at a temperature of 1.0 and the synthetic queries are generated by an LLM, retraining may yield results that slightly differ from the reported ones, whereas the accuracy evaluation itself is deterministic under greedy decoding.
\bibliography{iclr2027_conference}
\bibliographystyle{iclr2027_conference}

\clearpage
\appendix

\section*{Appendices}
\begingroup
    \hypersetup{
        linkcolor=magenta,
        citecolor=cyan,
        filecolor=magenta,      
        urlcolor=magenta,
        linktoc=all 
    }
  \setcounter{tocdepth}{3}
  \renewcommand{\contentsname}{Appendix Table of Contents}
  \tableofcontents
\endgroup

\clearpage
\section{Derivation of The Approximated Posterior Entropy}
\label{app-sec:proof}
We now derive an importance-reweighted expression for the posterior entropy $H(p_{\theta}(\cdot\mid q,z))$, which will later allow us to estimate (and optimize) this quantity using rollouts from the easier-to-sample prior distribution $p_{\theta}(\cdot\mid q)$ (\Eqref{eq:posterior_entropy_approx}). The key idea is to rewrite expectations under the conditional distribution $p_{\theta}(\cdot\mid q,z)$ as importance-weighted expectations under $p_{\theta}(\cdot\mid q)$ via Bayes' rule.
\begin{equation}
\label{eq:posterior_entropy_def}
H\!\left(p_{\theta}(\cdot\mid q,z)\right)=-\mathbb{E}_{o\sim p_{\theta}(\cdot\mid q,z)}\!\left[\log p_{\theta}(o\mid q,z)\right]=-\int p_{\theta}(o\mid q,z)\,\log p_{\theta}(o\mid q,z)\,\mathrm{d}o .
\end{equation}

By Bayes' rule, the conditional trajectory distribution can be expressed as
\begin{equation}
p_{\theta}(o\mid q,z)
=\frac{p_{\theta}(z\mid q,o)\,p_{\theta}(o\mid q)}{p_{\theta}(z\mid q)} ,
\label{eq:bayes_conditional}
\end{equation}
which implies the density ratio between the posterior and the prior:
\begin{equation}
\frac{p_{\theta}(o\mid q,z)}{p_{\theta}(o\mid q)}
=\frac{p_{\theta}(z\mid q,o)}{p_{\theta}(z\mid q)} .
\label{eq:is_ratio}
\end{equation}
In our setting, $p_{\theta}(z\mid q,o)$ is not directly observable; we therefore use the PRG-derived bounded signal $\mathcal{P}_{\mathrm{s}}$ as a monotone proxy for the ratio in \Eqref{eq:is_ratio}, i.e.,
\begin{equation}
    \frac{p_{\theta}(z\mid q,o)}{p_{\theta}(z\mid q)} \approx k\,\mathcal{P}_{\mathrm{s}},
\end{equation}
where $k$ denotes a proportional constant.

Substituting \Eqref{eq:bayes_conditional} into \Eqref{eq:posterior_entropy_def} and grouping terms yields an importance-sampling form:
\begin{equation}
\label{eq:posterior_entropy_is}
\begin{aligned}
&~H\!\left(p_{\theta}(\cdot\mid q,z)\right) \\
=& -\int k\mathcal{P}_{\mathrm{s}} p_{\theta}(o\mid q)\,
\log [k\mathcal{P}_{\mathrm{s}} p_{\theta}(o\mid q)]\,\mathrm{d}o \\
=& - k\mathbb{E}_{o\sim p_{\theta}(\cdot\mid q)}\!\left[
\mathcal{P}_{\mathrm{s}}\,\log p_{\theta}(o\mid q)\right] \\
& - k\mathbb{E}_{o\sim p_{\theta}(\cdot\mid q)}\!\left[
\mathcal{P}_{\mathrm{s}}\,\log \mathcal{P}_{\mathrm{s}} \right] \\
& - k\log k\mathbb{E}_{o\sim p_{\theta}(\cdot\mid q)}\!\left[
\mathcal{P}_{\mathrm{s}}\right].
\end{aligned}
\end{equation}
Therefore, up to a multiplicative constant and additive terms that do not affect the optimization
direction, we obtain
\begin{align}
H\!\left(p_{\theta}(\cdot\mid q,z)\right)&\propto -\mathbb{E}_{o\sim p_{\theta}(\cdot\mid q)}\!\left[\mathcal{P}_{\mathrm{s}}\log p_{\theta}(o\mid q)\right]\;+\;\text{const.}
\label{eq:entropy_mainterm}
\end{align}
The approximation in \Eqref{eq:entropy_mainterm} keeps the dominant term that couples the trajectory likelihood with the self-consistency weight $\mathcal{P}_{\mathrm{s}}$, while absorbing the remaining two terms into $\text{const.}$ for two practical reasons: (i) the factor $k$ is a normalization constant, so $k$ and $\log k$ only introduce a global scaling/shift; and (ii) $\mathcal{P}_{\mathrm{s}}$ is a bounded, saturated score (\cf \Eqref{eq.weight}), making the correction $\mathbb{E}[\mathcal{P}_{\mathrm{s}}\log \mathcal{P}_{\mathrm{s}}]$ comparatively small and mainly act as a regularizer on the score distribution rather than on the policy likelihood.

\clearpage
\section{Additional Implementation Details}
\label{app-sec:Implementation}
\subsection{Details of Constituent Strategies}
\label{app:Constituent}
\subsubsection{Reasoning Distillation (RD)} 
\label{app:RD}
RD~\citep{shridhar-etal-2023-distilling} trains the target policy $\pi_{\theta}$ to imitate the step-by-step reasoning trajectories from a stronger teacher model. Given the query $q$, the reasoning trajectory $r$ generated by the teacher, and the ground truth $y$, RD maximizes the length-normalized log-likelihood of the concatenated target sequence $s\!=\!(r,y)$ over a supervised corpus $\mathcal{D}$:
\begin{equation}
    \label{eq.rd}
    \mathcal{J}_{\mathrm{RD}}(\theta)\!=\!\mathbb{E}_{(q,s)\sim\mathcal{D}}\left[\frac{1}{|s|}\sum\nolimits_{t=1}^{|s|}\!\log \pi_{\theta}\!\left(s_t \!\mid\! q, s_{<t}\right)\right],
\end{equation}
where $s_t$ and $s_{<t}$ denote $t$-th token and the prefix of $s$ before position $t$, respectively. HyTuning instantiates $\mathcal{D}$ as $\mathcal{D}^{\mathrm{S}}\cup\mathcal{D}^{\mathrm{D}}$ in \Eqref{eq.final_rd}.

\subsubsection{Reinforcement Learning from Internal Feedback (RLIF)} 
\label{app:RLIF}
Unlike reinforcement learning with verifiable rewards (RLVR), which derives rewards from ground-truth answers, RLIF optimizes the policy with intrinsic signals of the LLM itself. We adopt INTUITOR~\citep{zhao2025learningreasonexternalrewards} as a representative RLIF algorithm, which uses the \emph{self-certainty} of each rollout as its reward within the GRPO framework~\citep{shao2024deepseekmathpushinglimitsmathematical}. For a rollout $o$ of query $q$, self-certainty averages the KL divergence between the uniform distribution $U$ over the vocabulary $V$ and the next-token distribution of the policy along $o$:
\begin{equation}
    \label{eq.self_certainty}
    c(o\mid q)=\frac{1}{|o|}\sum_{t=1}^{|o|}\mathbb{D}_{\mathrm{KL}}\!\big(U\,\|\,\pi_{\theta}(\cdot\mid q,o_{<t})\big)=-\frac{1}{|o|\cdot|V|}\sum_{t=1}^{|o|}\sum_{j=1}^{|V|}\log\!\big(|V|\cdot\pi_{\theta}(j\mid q,o_{<t})\big),
\end{equation}
where $|o|$ denotes the number of tokens in $o$. A higher $c(o\mid q)$ indicates a more concentrated next-token distribution and thus a more confident rollout. Since self-certainty does not depend on correctness, maximizing it alone may also reinforce confidence along erroneous trajectories~\citep{zhang2025freelunchrethinkinginternal}. Given a query corpus $\mathcal{D}$, INTUITOR samples $G$ rollouts $\{o_i\}_{i=1}^{G}$ from the old policy $\pi_{\theta_{\mathrm{old}}}$ for each query $q$ and maximizes the following GRPO-style objective:
\begin{equation}
    \label{eq.rlif}
    \begin{aligned}
    \mathcal{J}_{\mathrm{RLIF}}(\theta)&=\mathbb{E}_{q\sim\mathcal{D},\,\{o_i\}_{i=1}^{G}\sim\pi_{\theta_{\mathrm{old}}}(\cdot\mid q)}\Bigg[\frac{1}{G}\sum^{G}_{i=1}\!\frac{1}{|o_i|}\sum^{|o_i|}_{t=1}\Big(\min\big[w_{i,t}(\theta)A_{i,t},\,\\
    &\qquad\mathrm{clip}\big(w_{i,t}(\theta),1-\epsilon,1+\epsilon\big)A_{i,t}\big]-\beta\,\mathbb{D}_{\mathrm{KL}}(\pi_{\theta}\|\pi_{\mathrm{ref}})\Big)\Bigg],
    \end{aligned}
\end{equation}
where $w_{i,t}(\theta)=\frac{\pi_{\theta}(o_{i,t}\mid q,o_{i,<t})}{\pi_{\theta_{\mathrm{old}}}(o_{i,t}\mid q,o_{i,<t})}$ is the importance ratio, $\epsilon$ denotes the clipping range, and $\beta$ controls the KL penalty that regularizes the drift from the reference policy $\pi_{\mathrm{ref}}$ (\cf Table~\ref{tab:hytuning_hyperparams}). Following INTUITOR, the advantage of $o_i$ at each position $t$ is computed from the group-normalized self-certainty:
\begin{equation}
    \label{eq.advantage}
    A_{i,t}=\frac{c(o_i\mid q)-\mathrm{mean}\big(\{c(o_j\mid q)\}_{j=1}^{G}\big)}{\mathrm{std}\big(\{c(o_j\mid q)\}_{j=1}^{G}\big)},
\end{equation}
which guides $\pi_{\theta}$ toward rollouts that are more confident than their group average. 
Within HyTuning, the clipped surrogate of each rollout $o_i$ is multiplied by its own weight $\mathcal{P}_{\mathrm{s},i}$ from \Eqref{eq.weight} when the query comes from $\mathcal{D}^{\mathrm{S}}\cup\mathcal{D}^{\mathrm{D}}$, and remains unweighted when the query comes from $\mathcal{D}^{\mathrm{U}}$. 
Since \Eqref{eq:final_obj} conserves the total weight of each rollout, the intrinsic feedback withdrawn from a weakly grounded trajectory is reallocated to the anchoring gradient of the same query. Queries in $\mathcal{D}^{\mathrm{U}}$ lack such an anchor, so down-weighting them would merely shrink the learning signal of low-PRG trajectories without grounding them. Meanwhile, the anchored queries share the same policy with $\mathcal{D}^{\mathrm{U}}$, so their grounding gradients can also restrain the unconditional sharpening on $\mathcal{D}^{\mathrm{U}}$. Consistently, the self-certainty of HyTuning stays below that of INTUITOR throughout training (\cf Figure~\ref{fig:perf_conf_dynamics} (b)), and its most confident inferences remain the most reliable among all compared strategies (\cf Table~\ref{tab:faithfulness}). Extending PRG-based weighting to full unlabeled queries without anchors is left for future work.

\subsection{Details of Baselines}
\label{appendix:baselines}
To comprehensively validate the effectiveness of our HyTuning on both task performance and confidence faithfulness, we compare it against eight representative post-training strategies:
\begin{itemize}[leftmargin=*,topsep=0pt,parsep=0pt]
    \item[-] \emph{Supervised Fine-Tuning (SFT)}~\citep{ouyang2022traininglanguagemodelsfollow} represents the commonly-utilized process for adapting pre-trained LLMs to specific tasks by tuning them on a task-specific dataset with labeled examples rather than the reasoning traces. 
    \item[-] \emph{Reasoning Distillation (RD)}~\citep{yu2025primuspioneeringcollectionopensource, shridhar-etal-2023-distilling} typically collects reasoning trajectories from a stronger Large Reasoning Model and then distills the intermediate reasoning steps and final answers to guide LLMs towards the reasoning ability under the standard supervised learning objective ~\citep{guo2025deepseek}.
    \item[-] \emph{Group Relative Policy Optimization (GRPO)}~\citep{shao2024deepseekmathpushinglimitsmathematical} samples a group of responses per query and estimates the advantage of each response from its relative reward within this group. Its checkable reward is derived from the rule-based verification functions, which avoids the potential pitfalls of the learned reward models in RLHF~\citep{ouyang2022traininglanguagemodelsfollow}.
    \item[-] \emph{INTUITOR}~\citep{zhao2025learningreasonexternalrewards} replaces the pre-defined rewards in GRPO with its self-certainty, enabling unsupervised reinforcement learning.
    \item[-] \emph{Reinforcement Learning with Reference Probability Reward (RLPR)}~\citep{yu2025rlprextrapolatingrlvrgeneral} directly leverages LLM's intrinsic probability of generating a correct answer as the reward to incentivize reasoning in general domains without any external verifiers.
    \item[-] \emph{Hybrid Post-Training (HPT)}~\citep{lv2025unifiedviewlargelanguage} integrates SFT and RL as a unified post-training objective by dynamically adapting the mixing ratio between the SFT and RL losses, enabling the LLM to be intrinsically adaptive to its current status and the data with diverse complexities.
    \item[-] \emph{On-Policy Distillation (OPD)}~\citep{OPD} optimizes the student model on its self-generated trajectories with token-level supervision from a stronger teacher model, mitigating the training-inference distribution mismatch of conventional off-policy distillation.
    \item[-] \emph{On-Policy Self-Distillation (OPSD)}~\citep{OPSD} utilizes the same LLM as both student and teacher, where the student observes only the problem while the teacher additionally accesses verified reasoning traces to supervise the student's on-policy trajectories.
\end{itemize}

\subsection{Details of Evaluation Protocol}
\label{appendix:evaluation_protocol}
We take cybersecurity as our primary testbed, since it is one of the most representative high-stakes domains where confidently-wrong inferences may directly trigger system failures. Specifically, we evaluate all the post-trained LLMs on three publicly available cybersecurity-related benchmarks, namely ASBench~\citep{ma2025attackseqbenchbenchmarkinglargelanguage}, CSEBenchmark~\citep{wang2025digitalcybersecurityexpertfar}, and CyberMetric-500~\citep{tihanyi2024cybermetricbenchmarkdatasetbased}. Here, CSEBenchmark organizes expert-level cybersecurity knowledge into fine-grained knowledge points, CyberMetric covers a broad spectrum of cybersecurity sub-areas, and ASBench targets the reasoning over attack sequences described in Cyber Threat Intelligence reports. Together, they span threat analysis, vulnerability understanding, and attack sequence reasoning, enabling a systematic characterization of reasoning quality in high-stakes domains. Besides, we include MMLU~\citep{hendrycks2021measuringmassivemultitasklanguage} from the general domain to verify that the domain-specific gains of HyTuning do not come at the expense of its general capabilities. These four benchmarks are all formulated as multiple-choice questions, where a response is scored as correct only when the option extracted from its final answer matches the ground truth. To further verify that the contribution of our HyTuning is not confined to a single high-stakes domain, we re-instantiate the entire pipeline in the financial domain (\cf Section~\ref{sec.universality}) and evaluate it on FinQA~\citep{Finqa} and FinanceReasoning~\citep{Financereasoning} together with MMLU. FinQA requires numerical reasoning over hybrid tabular and textual evidence in financial reports, while FinanceReasoning further challenges LLMs with multi-step financial numerical reasoning across three difficulty levels. Unlike the above benchmarks, both financial benchmarks require free-form answers without candidate options, which we score by following their official answer-matching criteria. This setting rules out the possibility that the financial gains of HyTuning merely stem from exploiting answer-option patterns. All the evaluations are conducted with the \texttt{vLLM} inference engine under an identical prompt template across all the post-trained LLMs. We adopt greedy decoding for the accuracy evaluation to ensure deterministic and comparable outputs, and defer the sampling configuration of the uncertainty metrics to Appendix~\ref{appendix:metrics}.

\subsection{Details of Confidence Estimation}
\label{appendix:confidence_estimation}
The confidence faithfulness formalized in Section~\ref{sec:Preliminary} is defined over the ranking of the confidence score $c_i$, so its evaluation is only as reliable as the way $c_i$ is obtained. To obtain $c_i$, a seemingly direct option is to prompt the LLM to self-report its confidence score, yet such verbalized confidence tends to be overconfident~\citep{xiong2024llmsexpressuncertaintyempirical, groot2024overconfidence}. Its values mostly lie in the narrow range from 80\% to 100\% and typically take multiples of 5~\citep{xiong2024llmsexpressuncertaintyempirical}, which leads to extensive ties among instances and hinders the fine-grained confidence ranking required by Section~\ref{sec:Preliminary}. Moreover, although suitable prompting strategies can reduce its calibration error, verbalized confidence still struggles to distinguish correct answers from incorrect ones, with an average error rate of 47\% among the responses expressed with confidence~\citep{zhou2024relying}. Furthermore, prior studies attribute this overconfidence in part to post-training, whose preference data, reward models, and evaluation-driven objectives tend to reward confident answers over acknowledging uncertainty~\citep{zhou2024relying, leng2025taming, kalai2025language}. Since the compared post-training methods differ precisely in their optimization objectives, self-reported confidence may be confounded by how each objective reshapes the verbal expression, rather than solely reflecting the reliability of the inference.

We therefore derive the confidence score $c_i$ from two complementary uncertainty measures, both of which are computed from the output distribution of the LLM rather than from its verbal self-assessment. The first is \emph{Shifting Attention to Relevance (SAR)}~\citep{duan2024shifting}, which relies on multiple sampled generations and captures the uncertainty at the semantic level. Specifically, SAR measures the relevance of each token by the semantic change after removing it, and reweights the token-level and sentence-level uncertainty accordingly, which prevents semantically irrelevant tokens and generations from dominating the estimate. The second is \emph{Self-Certainty (SC)}~\citep{zhao2025learningreasonexternalrewards}, which is computed from the next-token distributions along a single generated response and captures the uncertainty at the token level. SC averages the KL divergence between the uniform distribution and the next-token distribution over all the generated tokens. For SAR, we take the negated uncertainty as the confidence score $c_i$, while for SC we directly adopt its value, so that a higher $c_i$ consistently indicates a more confident inference. Notably, although SC is adopted as the intrinsic reward by both INTUITOR and the RLIF branch of HyTuning, uniformly raising it does not improve the faithfulness metrics, since confidence faithfulness depends only on the ranking of $c_i$ (\cf Section~\ref{sec:Preliminary}). We thus treat SC as an auxiliary measure that examines whether the optimized intrinsic signal remains faithful to correctness after post-training, while SAR is not directly optimized by any compared method and serves as the primary independent reference. 

\subsection{Details of Evaluation Metrics}
\label{appendix:metrics}
We evaluate the proposed HyTuning and the post-training baselines along two dimensions, namely task performance and model confidence, whose consistency constitutes the \emph{confidence faithfulness} formalized in Section~\ref{sec:Preliminary}.

\textbf{Task performance.} We adopt \emph{Accuracy (Acc)} to measure task performance, which is defined as the proportion of instances whose final answer $\hat{y}_i$ matches the ground truth $y_i$. Here, $\hat{y}_i$ is obtained under greedy decoding (\cf Appendix~\ref{appendix:evaluation_protocol}).

\textbf{Confidence faithfulness.} 
Following Section~\ref{sec:Preliminary}, we rank all the instances in descending order of their confidence scores $c_i$ (\cf Appendix~\ref{appendix:confidence_estimation}), and quantify confidence faithfulness with three complementary metrics that focus on the reliability of the high-confidence region rather than the entire confidence range. \emph{Selective Accuracy (SelAcc@$\kappa$)} denotes the accuracy over the most confident $\kappa$ fraction of instances~\citep{farquhar2023semanticentropy}, which directly instantiates the criterion in Section~\ref{sec:Preliminary}. A confidence-faithful LLM should exhibit an increasing SelAcc@$\kappa$ as $\kappa$ shrinks. \emph{Generalized Risk (GR@$\kappa$)} represents the proportion of all instances that fall into the most confident $\kappa$ fraction yet are answered incorrectly~\citep{traub2024overcoming}, where a lower value is better. Although SelAcc@$\kappa$ already measures the reliability of the retained instances, GR@$\kappa$ further counts the confidently-wrong inferences over all the queries. It directly reflects how often a deployed LLM delivers confidently-wrong inferences to practitioners and remains comparable across different $\kappa$. \emph{High-confidence Area Under the Risk-Coverage curve (hAURC)} restricts the standard AURC~\citep{geifman2018bias} to the most confident $\kappa_h$ fraction, which averages the selective risk (\ie the error rate among the retained instances) over all the coverage levels up to $\kappa_h$. Different from SelAcc@$\kappa$ and GR@$\kappa$ at a single operating point, hAURC summarizes the reliability of the entire high-confidence region, where a lower value is better. In summary, the three metrics characterize confidence faithfulness from complementary perspectives, namely the reliability at a given coverage (SelAcc@$\kappa$), the exposure to confidently-wrong inferences (GR@$\kappa$), and the overall risk of the high-confidence region (hAURC). Consistent with the asymmetric criterion in Section~\ref{sec:Preliminary}, they only evaluate on the high-confidence region and impose no requirement on the low-confidence one. Notably, all of them only depend on the ranking order of $c_i$, making them invariant to any monotone rescaling of confidence scores and enabling fair comparisons across post-training strategies and uncertainty measures.

\subsection{Implementation Details of Baselines}
\label{appendix:ID_baselines}
To ensure the fair comparison, we implement SFT, RD, GRPO, INTUITOR, OPD and OPSD with the same \texttt{verl}~\citep{sheng2025hybridflow} codebase as HyTuning. For RLPR and HPT, we directly adopt their official codebases to avoid potential discrepancies introduced by re-implementation. All the baselines are trained for a single epoch using the AdamW optimizer with $\beta_1\!=\!0.9$ and $\beta_2\!=\!0.999$, a global batch size of 64, and a micro-batch size of 1 per GPU. Under partial supervision, we reduce the global batch size of the supervised strategies to 2, so that their 100 supervised samples yield a comparable number of optimization steps to the unsupervised strategies and our HyTuning. The learning rate is fixed to $3\times10^{-6}$ for all the baselines. For SFT and RD, the maximum sequence length is set to 8,192 tokens. For the RL-based baselines (\ie GRPO, INTUITOR, RLPR, and HPT), the maximum prompt and completion lengths are set to 1,024 and 4,096 tokens, with a sampling temperature of 1.0 and 8 rollouts per prompt. SFT, RD, GRPO, and INTUITOR adopt a cosine learning rate schedule with a warmup ratio of 0.1, and GRPO and INTUITOR further apply a KL penalty with a coefficient of 0.005. In contrast, RLPR and HPT follow their official configurations with a constant learning rate and without warmup or KL penalty, where we only adapt their data formats and prompt templates to our setting. Regarding the distillation-based baselines, OPD adopts the base version of Qwen2.5-7B as the teacher model to provide token-level supervision on the on-policy trajectories of the student. OPSD instead employs the same LLM as both the student and the teacher, where the teacher is additionally conditioned on the verified reasoning traces as privileged information.

\subsection{Implementation Details of HyTuning}
\label{appendix:ID_HyTuning}

\begin{wraptable}{r}{0.46\textwidth}
\vspace{-0.45cm}
\centering
\caption{Hyperparameters used for HyTuning.}
\vspace{-0.3cm}
\label{tab:hytuning_hyperparams}
\begin{tabular}{l c}
\toprule
\textbf{Hyperparameter} & \textbf{Value} \\
\midrule
Optimizer & AdamW \\
$\beta_1$ & 0.9 \\
$\beta_2$ & 0.999 \\
Learning rate & $3 \times 10^{-6}$ \\
Learning rate scheduler & Cosine \\
Warmup ratio & 0.1 \\
Global batch size & 64 \\
Micro-batch size (per GPU) & 1 \\
Number of epochs & 1 \\
Max prompt length & 1024 \\
Max completion length & 4096 \\
Sampling temperature & 1.0 \\
Number of rollouts & 8 \\
KL penalty coefficient & 0.005 \\
\bottomrule
\vspace{-0.3cm}
\end{tabular}
\end{wraptable}
We adopt Qwen2.5-3B as the backbone LLM in our primary experiments. 
Under partial supervision, HyTuning is trained on 100 supervised reasoning traces and 3,000 unlabeled queries, whose sources and construction are detailed in Appendix~\ref{appendix:training_corpus}, while the fully supervised setting uses the complete supervised corpus. Our implementation builds upon the codebase of INTUITOR~\citep{zhao2025learningreasonexternalrewards}, which is developed on the \texttt{verl}~\citep{sheng2025hybridflow} framework. To ensure a fair comparison, HyTuning shares the same optimizer, learning rate schedule, rollout configuration, and KL penalty with GRPO and INTUITOR in Appendix~\ref{appendix:ID_baselines}. For the adaptive PRG weight in Equation~\ref{eq.weight}, we set $\alpha$=0.5 and $\tau$=0.8. Unless otherwise specified, all experiments share the same training configuration and hyperparameters. The detailed hyperparameters utilized for our HyTuning are summarized in Table~\ref{tab:hytuning_hyperparams}.

\subsection{Computational Resources}
\label{appendix:Computational_Resource}
The experiments on the 3B backbones (\ie Qwen2.5-3B, LLaMA3.2-3B, and Ministral3-3B) and Qwen2.5-7B are conducted on NVIDIA A100 GPU with 80~GB of memory, and the experiments on Qwen2.5-14B are conducted on NVIDIA H200 GPU with 141~GB of memory to accommodate its larger memory footprint. For each backbone, the same hardware is used for both training and evaluation.

\subsection{Training Corpus Collection}
\label{appendix:training_corpus}
We construct the post-training corpus in cybersecurity domain upon Primus-Reasoning~\citep{yu2025primuspioneeringcollectionopensource}, which originally contains 4,891 complete reasoning trajectories generated by \texttt{GPT-o1-preview} and \texttt{DeepSeek-R1} on cybersecurity tasks. Although rejection sampling has already ensured the answer correctness of these reasoning trajectories, we further apply a two-stage filtering pipeline to eliminate the residual noise in their language and reasoning. Specifically, we first leverage FastText~\citep{joulin2017bag} to discard all the non-English instances. Then we discard the instances whose trajectory contains more than 10 transitional words (\eg ``However'' and ``Wait''), since such words indicate a departure of the current reasoning step from the preceding analysis~\citep{zhang2025freelunchrethinkinginternal}, and their frequent occurrence implies that the intermediate conclusions are repeatedly overturned rather than progressively reinforced, the very pattern that PRG captures (\cf Section~\ref{sec:Preliminary}). Meanwhile, this threshold still allows moderate transitions for necessary exploration. After filtering, we retain 3,112 out of the original 4,891 instances as the supervised corpus, each of which consists of a query, a reasoning trace, and a ground-truth answer.

Based on this corpus, we construct two supervision regimes for the performance comparison in Section~\ref{sec.PerformanceComparison}. Under full supervision, the supervised post-training strategies are trained on all the 3,112 instances, while the unsupervised post-training strategies are trained on the same queries with their reasoning traces and answers removed. Under partial supervision, we simulate the data scarcity in high-stakes domains by randomly selecting 100 instances as the only accessible data, on which the supervised post-training strategies are trained. Since unsupervised post-training strategies does not require ground truth, we further employ \texttt{Qwen2.5-32B-Instruct} to generate 3,000 synthetic queries by taking the 100 selected queries as seeds. Only the generated queries are retained for these unsupervised post-training strategies. This design ensures that the partial regime accesses no real instance beyond the 100 seeds. Within each regime, HyTuning jointly consumes the supervised and unsupervised data. Beyond the aforementioned two regimes, we further randomly select 500 and 1,000 instances from the supervised corpus for the scaling potential analysis in Section~\ref{sec.scaling_potential}, where the ``All Corpus'' setting corresponds to all the 3,112 instances and the selection criterion of the ``Top-100 Samples'' setting is detailed in Appendix~\ref{app:top100_training_corpus}. For the universality analysis in Section~\ref{sec.universality}, we construct the post-training corpus in the financial domain following the same procedure. Specifically, we draw the supervised instances from the FinCoT dataset~\citep{nitarach2025fincot}, which pairs financial questions with reasoning traces generated by \texttt{GPT-4o}, and likewise obtain 100 supervised instances and 3,000 synthetic queries under partial supervision.

\subsection{Selection Criterion of the Top-100 Supervised Traces}
\label{app:top100_training_corpus}
We construct the Top-100 subset to verify whether the quality of supervised traces can substitute for their volume, holding the volume at 100 samples and selecting the traces by quality rather than at random. Considering that the quality of a given reasoning trace in a high-stakes domain cannot be fully captured by surface statistics, we have recruited three cybersecurity experts to specify what a competent answer to the domain-specific question has to demonstrate and turned their specification into a judging prompt in Box~\ref{prompt:data_filter_quality_eval}. This criterion evaluates every instance on (1) whether the question is grounded in cybersecurity, vulnerability analysis or CTI understanding and connects to pre-established knowledge bases such as CWE, CVSS and MITRE ATT\&CK, (2) whether the reasoning trace is factually correct, complete and technically sound, and (3) whether both are clearly and concisely phrased. We apply this prompt to the supervised corpus in Appendix~\ref{appendix:training_corpus} with \texttt{GPT-5} as the judge, retain only the traces rated high on all three aspects, and draw 100 of them uniformly at random. Since both settings draw 100 instances uniformly from their respective candidates and share the same 3,000 synthetic queries for unsupervised post-training, the ``Top-100 setting'' differs from the main experiments only in the quality stratum of the supervised data.

\clearpage
\section{Additional Experiments}
\label{app-sec:Experiments}

\subsection{PRG as a One-Sided Reliability Indicator}
\label{app:prg-screen}
\begin{wraptable}{r}{0.7\textwidth}
\vspace{-0.4cm}
\centering
\caption{Human evaluation of 100 reasoning traces across five PRG bins on CyberMetric. CS and DC are averaged over four experts, CS=0 and DC=0 report the fraction of traces receiving a zero from at least one expert on the corresponding criterion.}\label{tab:prg_human}
\vspace{-0.2cm}
\resizebox{\linewidth}{!}{
\begin{tabular}{lccccc}
\toprule
PRG bin & B1 (lowest) & B2 & B3 & B4 & B5 (highest) \\
\midrule
CS ($\uparrow$) & 1.513 & 1.750 & 1.788 & 1.838 & 1.875 \\
DC ($\uparrow$) & 1.588 & 1.863 & 1.913 & 1.875 & 1.913 \\
\midrule
CS=0 ($\downarrow$) & 30\% & 10\% & 10\% & 5\% & 0\% \\
DC=0 ($\downarrow$) & 25\% & 5\% & 5\% & 5\% & 0\% \\
\bottomrule
\end{tabular}}
\vspace{-0.4cm}
\end{wraptable}
Confidence accumulation is a necessary rather than a sufficient property of a well-grounded reasoning trajectory, and PRG inherits this asymmetry by accumulating only the positive increments of the LLM's confidence in the final answer (\cf \Eqref{eq:prg}). Therefore, a near-zero PRG marks a weakly grounded trajectory regardless of whether its answer is correct. The ReLU is essential to this design. \textbf{Without it, the signed increments would telescope to $\log p_{\pi_\theta}(\hat{y}\mid q,o_{\le T})-\log p_{\pi_\theta}(\hat{y}\mid q,o_{\le 0})$ and discard how confidence evolves along the trace, while exploratory steps that temporarily lower confidence would also be penalized}. However, the tolerance of temporary drops means that a high PRG does not guarantee steady convergence, since the rebounds of an oscillating trajectory are also counted as positive increment. To verify this one-sided role, we randomly sample 100 samples from CyberMetric together with their reasoning traces generated by HyTuning (Qwen2.5-3B), rank them by PRG and split them into five equal-sized bins from B1 (lowest) to B5 (highest). Then we recruit four human experts to independently evaluate each trace without access to its PRG score. Each expert rates every trace on two criteria using a 0--2 scale. Specifically, \emph{Conclusion Support} (CS) assesses whether the final answer follows from the preceding analysis, ranging from 0 when the analysis contradicts or fails to support the conclusion to 2 when the conclusion clearly follows from it. \emph{Directional Consistency} (DC) examines whether confidence develops in a consistent direction, where 2 indicates that confidence consistently moves toward the final answer and 0 indicates clear reversals or self-contradictions.

From Table~\ref{tab:prg_human}, we can observe that: 
(1) Weakly-grounded reasoning trajectories concentrate in the low-PRG regime. B1 obtains the lowest CS and DC for each of the four experts individually. Compared with the remaining 80 traces, B1 scores 0.300 lower on CS and 0.303 lower on DC, and both gaps are significant under a one-sided permutation test ($p$=0.013 and $p$=0.006). These gaps mainly stem from severe failures, where 30\% and 25\% of B1 traces receive zero from at least one expert on CS and DC, against only 6.25\% and 3.75\% of the remaining traces (one-sided Fisher's exact test with $p$=0.008 for each criteria). This supports the role of PRG in HyTuning, where a low-PRG trajectory receives weaker intrinsic sharpening as prescribed by Principle~1, while the verified target of the same query receives a larger RD weight as prescribed by Principle~2.
(2) Beyond the low-PRG regime, PRG no longer distinguishes better-grounded reasoning trajectories. Across B2 to B5, CS rises by 0.125, while DC fluctuates between 1.863 and 1.913 without a monotone trend. The flat DC trend beyond B2 reflects a deliberate trade-off in PRG. The ReLU ignores temporary confidence drops so that trajectories revisiting and correcting an intermediate step are not penalized, at the cost of not distinguishing such benign revisions from harmful reversals among high-PRG trajectories. This observation supports the saturating map in \Eqref{eq.weight}, which preserves discriminability in the low-PRG regime and assigns nearly uniform weights once PRG saturates, so that marginal PRG differences among adequate trajectories barely change the balance between RLIF and RD.
(3) A low PRG indicates an elevated risk of weak grounding rather than a definite reasoning failure. Even in B1 that obtains the lowest average scores, only 30\% and 25\% of traces are rated zero by any expert on CS and DC, respectively. A hard filter that discards low-PRG trajectories would thus also remove sound reasoning. This supports acting on PRG through graded weighting rather than hard filtering. HyTuning down-weights rather than discards the intrinsic feedback of low-PRG trajectories, and assigns a correspondingly larger RD weight to the verified target of the same query, so that the optimization relies more heavily on the verified reasoning traces.

Finally, we verify the reliability of the expert ratings and the robustness of these findings to analysis choices. Regarding inter-expert alignment, the four experts reach an ordinal Krippendorff's $\alpha$ of 0.59 on CS and 0.51 on DC. Since most HyTuning traces are rated as well-grounded (81\% and 88\% of ratings take the maximum score), agreement by chance is already high, which lowers the chance-corrected $\alpha$. Moreover, the disagreements are mostly confined to adjacent levels, where the four ratings differ by at most one point on 96\% and 97\% of traces. Since our analysis relies on scores averaged over four experts, we further assess the reliability of these averaged scores, obtaining an ICC(2,$k$) of 0.89 on both criteria. Regarding robustness, the discovery about ``Weakly-grounded reasoning trajectories concentrate in the low-PRG regime'' persists when any single expert is left out ($p\le0.03$) and when trace length is controlled as a covariate to rule out that PRG merely reflects trace length ($p\le$0.03). It also largely holds when the low-PRG group is defined as the bottom 15\% to 50\% of traces in steps of 5\%, where the gap remains significant at every cutoff on DC ($p\le$0.042) and at all but the 30\% and 35\% cutoffs on CS ($p$=0.054 and $p$=0.058). Excluding the expert with the lowest agreement with the others further raises $\alpha$ from 0.59 to 0.61 on CS and from 0.51 to 0.68 on DC. Under this setting, B1 still scores 0.300 lower on CS and 0.308 lower on DC than the remaining 80 traces, and both gaps remain significant under the one-sided permutation test ($p$=0.014 and $p$=0.004).

\begin{table*}[!t]
\centering
\footnotesize
\caption{Performance comparison of ablations on cybersecurity-related and general benchmarks.}
\vspace{-0.2cm}
\setlength{\tabcolsep}{4pt}
\label{tab.comprehensive_ablation}
\begin{tabular}{c|ccc|c|c}
\toprule
\multirow{2}{*}{\textbf{Variants}} & \multicolumn{3}{c|}{\textbf{Domain-specific Benchmarks}} & \textbf{Gen. Bench.} & \multirow{2}{*}{\textbf{Average}} \\
& \textbf{ASB} & \textbf{CSEB} & \textbf{CyberM} & \textbf{MMLU} & \\ \midrule
\textbf{HyTuning} & \textbf{0.489} & \textbf{0.391} & \textbf{0.732} & \textbf{0.552} &  \textbf{0.541} \\ \midrule
\textbf{HyTuning - RLIF} & 0.460 & 0.328 & 0.666 & 0.457 & 0.478 \\
\textbf{HyTuning - RD} & 0.476 & 0.321 & 0.700 & 0.476 & 0.494 \\
\textbf{HyTuning - AR} & 0.463 & 0.309 & 0.716 & 0.464 & 0.489 \\
\textbf{HyTuning - PRG} & 0.464 & 0.335 & 0.712 & \textbf{0.552} & 0.516 \\
\textbf{Continual Tuning (RD $\rightarrow$ RLIF)} & 0.452 & 0.327 & 0.664 & 0.463 & 0.477 \\
\textbf{Continual Tuning (RLIF $\rightarrow$ RD)} & 0.472 & 0.326 & 0.714 & 0.475 & 0.498 \\
\bottomrule
\end{tabular}
\vspace{-0.5cm}
\end{table*}

\subsection{Comprehensive Results of Surgical Analysis}
\label{appendix:com_results}
Due to the page limitation, we report the comprehensive experimental results of the surgical analysis in Table~\ref{tab.comprehensive_ablation}, where each ``HyTuning - X'' variant corresponds to ``Ours-X'' in Figure~\ref{fig:Ablation_Study}. Here, ``HyTuning - RLIF'' and ``HyTuning - RD'' remove one of the constituent strategies and therefore reduce to RD and INTUITOR alone, respectively. ``HyTuning - AR'' removes the adaptive reweighting and combines the two strategies with a static equal weight. ``HyTuning - PRG'' replaces PRG with self-certainty as the adaptive signal while keeping the remaining design unchanged. Beyond the hybrid post-training paradigm, we further consider two continual tuning variants that apply the two strategies sequentially, where ``Continual Tuning (RD$\rightarrow$RLIF)'' first trains with RD and then with RLIF, whereas ``Continual Tuning (RLIF$\rightarrow$RD)'' reverses this order. Each stage follows the configuration in Appendix~\ref{appendix:ID_HyTuning}.

As illustrated in Table~\ref{tab.comprehensive_ablation}, we have the following discoveries: 
(1) HyTuning obtains the best or tied-best performance on every benchmark, and removing either constituent strategy causes a clear performance degradation. The average accuracy drops from 0.541 to 0.478 for ``HyTuning - RLIF'' and 0.494 for ``HyTuning - RD'', indicating that RD and RLIF are complementary rather than redundant in our HyTuning.
(2) ``HyTuning - AR'' even falls below ``HyTuning - RD'', which indicates that fusing the two constituent strategies with a static weight degrades the stronger constituent strategy, and its accuracy on CSEB is the lowest among all variants and even below the backbone. This observation is consistent with our motivation that indiscriminate fusion amplifies erroneous updates. It further implies that the benefit of hybrid post-training arises from how the two signals are weighted rather than from their mere coexistence.
(3) ``HyTuning - PRG'' is the strongest ablation variant on average and exhibits comparable performance with HyTuning on MMLU, yet it underperforms HyTuning on all three domain-specific benchmarks. Since ``HyTuning - PRG'' and ``HyTuning'' share the identical training pipeline and differ only in the signal that drives the adaptive weight, this gap isolates the contribution of PRG. We attribute it to the over-self-certainty, which is also the reward maximized by the RLIF branch and thus assigns large weights to confidently-wrong trajectories, whereas PRG examines how confidence evolves along the reasoning trace. Therefore, PRG strengthens domain-specific reasoning without sacrificing general capability.
(4) Each continual tuning variant stays close to the strategy applied in its first stage and this paradigm is sensitive to the order of its two stages, where ``Continual Tuning (RLIF$\rightarrow$RD)'' reaches 0.498 whereas ``Continual Tuning (RD$\rightarrow$RLIF)'' drops to 0.477 and even falls below the static equal-weight variant. Neither of them approaches HyTuning, since a predetermined schedule cannot respond to the reliability of individual reasoning trajectories.
Overall, these observations demonstrate that every component within HyTuning is indispensable to its effectiveness, which complements its confidence faithfulness reported in Section~\ref{sec:conf_faith} and Appendix~\ref{appendix:faithfulness_region}.

\subsection{Other Training Dynamics Analysis}
\label{appendix:training_dynamic}
Beyond the performance and self-certainty dynamics reported in Section~\ref{sec:training_dynamic}, we further analyze how the PRG score and the transitional word frequency evolve over training for the same five settings, namely RD (Full), INTUITOR (Partial), INTUITOR (Full), HyTuning (Partial), and HyTuning (Full). Here, ``Partial'' and ``Full'' denote training on the 100-instance supervised corpus adopted in our main experiments and on the complete corpus, respectively. RD is reported under the complete corpus alone, where it serves as the fully supervised reference for both metrics. We analyze the two quantities in Appendix~\ref{app.prg_dynamic} and Appendix~\ref{app.trans_dynamic}, respectively.

\begin{wrapfigure}{r}{0.64\textwidth}
    \vspace{-0.4cm}
    \centering
    \includegraphics[width=1.0\linewidth]{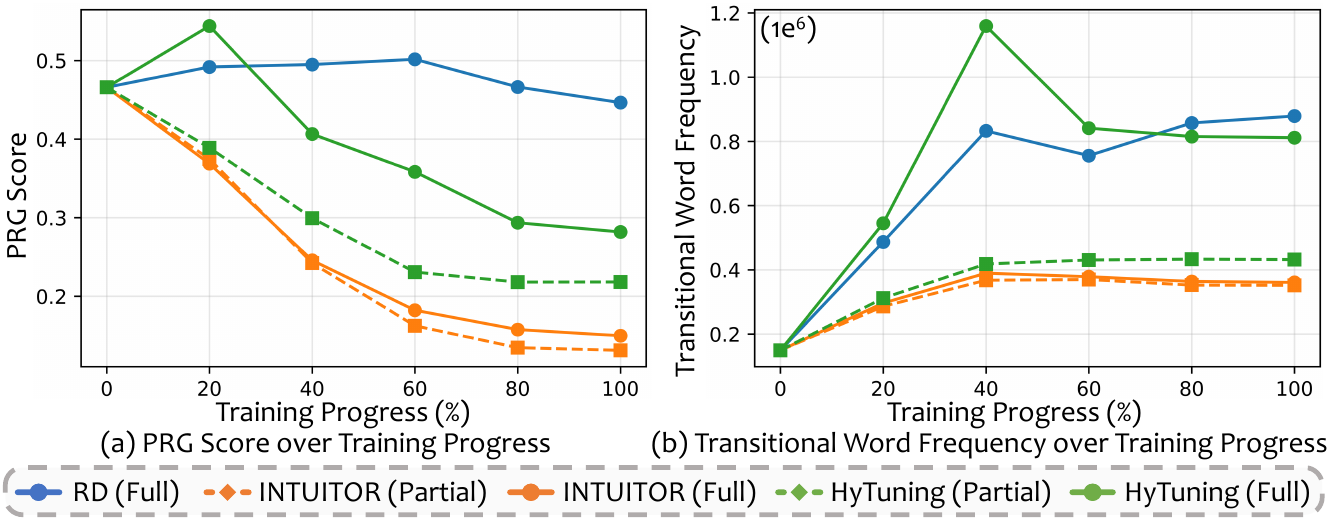}
    \vspace{-0.4cm}
    \caption{Comparison of PRG score and transitional word frequency over training, where Partial and Full denote the 100-sample and the complete supervised corpus.}
    \vspace{-0.4cm}
    \label{fig:PRG_Tran_dynamics}
\end{wrapfigure}
\subsubsection{Evolution of Progressive Reasoning Gain over Training}
\label{app.prg_dynamic}
PRG serves as the adaptive signal within HyTuning that mediates between RD and RLIF rather than as an optimization target. A change in its value therefore does not indicate a change in performance, but a shift in how progressively the model accumulates confidence over its own reasoning trace. As shown in Figure~\ref{fig:PRG_Tran_dynamics}~(a), RD maintains the most progressive accumulation throughout the whole training process, since it imitates distilled traces whose intermediate steps are elaborately constructed to build toward the answer. INTUITOR sits at the opposite extreme under both training corpora, where maximizing self-certainty saturates confidence early in the reasoning trace and leaves few increments for the remaining steps. HyTuning consistently lies between these two extremes and never collapses onto either of them. Since $\mathcal{P}_\mathrm{s}$ increases monotonically with PRG in \Eqref{eq.weight} and is evaluated per trajectory, it acts as a selective router on queries with verified targets, which reserves the intrinsic reward for a trajectory whose confidence accumulates progressively and falls back on the verified target otherwise, so that confidence grows mainly where the reasoning supports it. Therefore, the decline in Figure~\ref{fig:PRG_Tran_dynamics}~(a) reflects a changing mixture across trajectories rather than a collapse onto RD, which is precisely the selective sharpening prescribed by \textbf{Principle~1}.
 
\subsubsection{Evolution of Transitional Word Frequency over Training}
\label{app.trans_dynamic}
Unlike the self-certainty in Section~\ref{sec:training_dynamic} and the PRG score in Appendix~\ref{app.prg_dynamic} that are both derived from the output distribution of the LLM, the transitional word frequency is counted directly from the generated text and therefore offers an independent behavioral readout. 
Following \citet{zhang2025freelunchrethinkinginternal}, we count discourse connectives and transitional cues such as ``however'' and ``wait'' within the LLM output, which indicate that the LLM is still entertaining alternatives and which also mark the ``aha moment'' reported in \citet{guo2025deepseek}. A higher frequency thus reflects a more deliberative reasoning process, whereas a lower one reflects a more committed process. Notably, this readout is not monotone in reasoning quality. Excessive transitions indicate repeatedly overturned conclusions, which is why Appendix~\ref{appendix:training_corpus} discards training traces containing more than 10 of them. We therefore interpret the frequency relative to RD, whose training traces already satisfy this cap. We report the aggregate count rather than a length-normalized rate, since the amount of deliberative language that the LLM emits is itself part of the behavior under study and a model that reaches its answer sooner produces both a shorter trace and fewer cues within it. 
Every variant optimized with the intrinsic reward eventually flattens in Figure~\ref{fig:PRG_Tran_dynamics}~(b) while its self-certainty in Figure~\ref{fig:perf_conf_dynamics}~(b) continues to rise, so the flattening alone does not distinguish them. 
What helps us to distinguish them is where each one flattens and how much certainty it reaches by the end of training. 
Both INTUITOR variants converge to the highest self-certainty and the lowest transitional word frequency, whereas both HyTuning variants converge to a lower self-certainty and a higher transitional word frequency. HyTuning (Full) settling close to RD, which generates the most transitional words among all five variants.
Alongside this trend in transitional word frequency, HyTuning attains higher accuracy than both INTUITOR variants in Figure~\ref{fig:perf_conf_dynamics}~(a), and retains this advantage from only 100 supervised instances as shown in Section~\ref{sec.scaling_potential}, so it inherits the scalability of RLIF without inheriting its drift away from confidence faithfulness.

\begin{wrapfigure}{r}{0.35\textwidth}
\vspace{-0.45CM}
    \centering
    \begin{minipage}[t]{0.35\textwidth}
        \centering
        \includegraphics[width=0.98\linewidth]{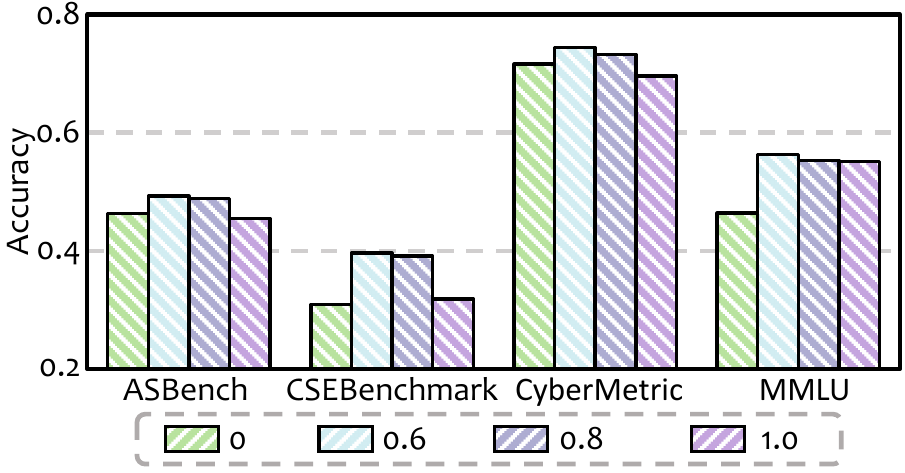}
        \vspace{-0.3cm}
        \caption{Performance comparison of our HyTuning equipped with different temperature $\tau$.}
        \label{fig:parameter_temp}
    \end{minipage}\hfill
    \begin{minipage}[t]{0.35\textwidth}
        \centering
        \includegraphics[width=0.98\linewidth]{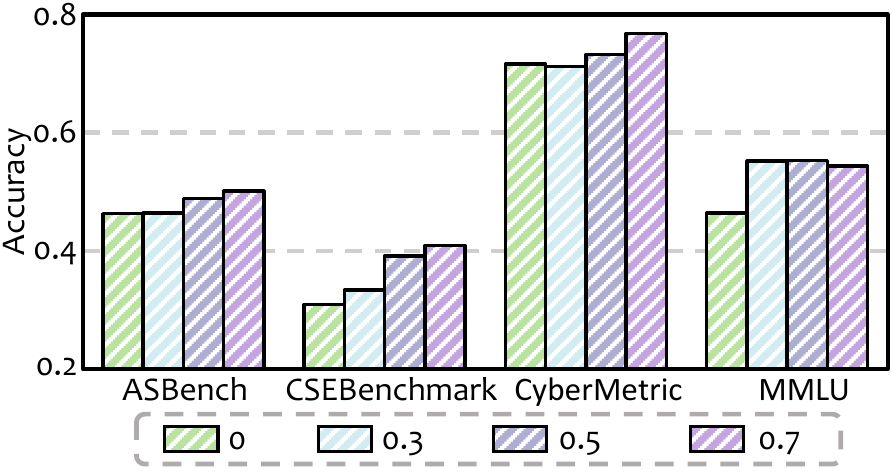}
        \vspace{-0.3cm}
        \caption{Performance comparison of our HyTuning equipped with different scaling factor $\alpha$.}
        \vspace{-0.5cm}
        \label{fig:hyperparam_alpha}
    \end{minipage}
\end{wrapfigure}
\subsection{Hyper-Parameter Sensitivity Analysis}
To evaluate the robustness of HyTuning, we investigate its sensitivity to the stemperature $\tau$ and the scaling factor $\alpha$ in \Eqref{eq.weight}, which respectively control the smoothness and the magnitude of the PRG-based weighting mechanism. Setting either one to zero removes the adaptivity of $\mathcal{P}_{\mathrm{s}}$, and we implement this setting as the equal-weight combination of RD and RLIF. Therefore, the leftmost bars of Figure~\ref{fig:parameter_temp} and Figure~\ref{fig:hyperparam_alpha} coincide on every benchmark. Notably, HyTuning targets a broad range of high-stakes domains. To preserve its generality, we do not tune hyper-parameters for individual domains or benchmarks, and instead adopt a single shared configuration ($\tau$=0.8, $\alpha$=0.5) across all experiments.

The temperature $\tau$ determines how quickly $\mathcal{P}_{\mathrm{s}}$ saturates at its upper bound $\alpha$. Since the ReLU in \Eqref{eq:prg} restricts $\mathcal{P}$ to $[0,+\infty)$, an overly large $\tau$ tends to push most trajectories toward the same weight, which weakens the selectivity of PRG-based weighting mechanism. As shown in Figure~\ref{fig:parameter_temp}, accuracy on all four benchmarks rises from $\tau$=0 to $\tau$=0.6 and remains comparable at $\tau$=0.8. However, accuracy declines at $\tau$=1.0 with the largest drop on CSEBenchmark, which is consistent with the saturation effect above. The scaling factor $\alpha$ sets the maximum loss weight that PRG weighting can allocate to RLIF and RD. Figure~\ref{fig:hyperparam_alpha} shows that a larger $\alpha$ generally benefits the domain-specific benchmarks, where $\alpha=0.7$ achieves the highest accuracy on them. In contrast, accuracy on MMLU remains largely stable for $\alpha \geq 0.3$ and slightly decreases at $\alpha=0.7$. This suggests that a larger $\alpha$ favors domain-specific gains at a small cost of general capability. We therefore adopt $\alpha=0.5$ as a moderate trade-off. Overall, the shared configuration is not the per-benchmark optimum for either hyper-parameter. For instance, $\tau=0.6$ yields slightly higher accuracy on all four benchmarks, and $\alpha=0.7$ yields higher accuracy on the three domain-specific benchmarks. The reported results are thus likely a conservative estimate of HyTuning, and fine-grained domain-specific tuning may further improve its performance.

\begin{wraptable}{r}{0.72\textwidth}
\vspace{-0.45cm}
\centering
\footnotesize
\caption{Robustness of HyTuning to the internal reward within the RLIF branch. Each block pairs a baseline with the HyTuning variant that uses the identical reward, where the better result within a block is in bold.}
\vspace{-0.2cm}
\setlength{\tabcolsep}{3pt}
\label{tab.reward}
\begin{tabular}{C{3.6cm}|C{0.8cm}C{0.8cm}C{1.0cm}|C{1.66cm}|C{0.7cm}}
\toprule
\multirow{2}{*}{\textbf{Models}} & \multicolumn{3}{c|}{\textbf{Domain-specific Bench.}} & \textbf{Gen. Bench.} & \multirow{2}{*}{\textbf{Ave.}} \\
 & ASB & CSEB & CyberM & MMLU & \\ \midrule
Qwen2.5-3B & 0.269 & 0.318 & 0.556 & 0.449 & 0.398 \\ \midrule
INTUITOR & 0.476 & 0.321 & 0.700 & 0.476 & 0.494 \\
HyTuning w/ INTUITOR & \textbf{0.489} & \textbf{0.391} & \textbf{0.732} & \textbf{0.552} & \textbf{0.541} \\ \midrule
EM-RL-Token & 0.281 & 0.340 & 0.622 & 0.457 & 0.425 \\
HyTuning w/ EM-RL-Token & \textbf{0.501} & \textbf{0.452} & \textbf{0.786} & \textbf{0.531} & \textbf{0.568} \\ \bottomrule
\end{tabular}
\vspace{-0.3cm}
\end{wraptable}

\subsection{Robustness to the Internal Reward Estimator}
\label{app.reward}
The RLIF branch of HyTuning adopts self-certainty as its intrinsic reward, following the official implementation of INTUITOR~\citep{zhao2025learningreasonexternalrewards}. A natural concern is whether the exhibited performance improvements depend on this particular reward rather than on the hybrid post-training mechanism itself. To examine this, we replace the reward of the RLIF branch with the token-level entropy minimization of EM-RL~\citep{EM-RL}, and we refer to this variant as \textbf{EM-RL-Token}. All post-training corpora, hyper-parameters, and evaluation protocols follow the same setup of Appendix~\ref{appendix:ID_HyTuning}. As illustrated in Table~\ref{tab.reward}, we observe that HyTuning surpasses its RLIF implementation on every benchmark under both intrinsic rewards, which indicates that the improvement stems from the hybrid mechanism rather than from a particular confidence estimator. We further observe that the two variants reach comparable performance (0.541 and 0.568), although their RLIF baselines differ considerably (0.494 and 0.425). Therefore, HyTuning depends far less on the strength of the underlying intrinsic reward than the RLIF variants do. Beyond this consistency, the two intrinsic rewards exhibit complementary tendencies in our implementation. EM-RL-Token drives HyTuning toward stronger domain specialization (0.580 against 0.537 on domain-specific benchmarks), whereas self-certainty better preserves general capability (0.552 against 0.531 on MMLU). We adopt self-certainty throughout this work for this balance and regard the token-level entropy variant as an effective instantiation of the RLIF branch.

\begin{table}[t]
\centering
\small
\caption{Confidence faithfulness on CyberMetric-500 under SAR and SC within three high-confidence regions ($\kappa\in\{5\%,10\%,20\%\}$). For SelAcc@$\kappa$ and hAURC, cells are shaded by rank across the three regions within each row: \colorbox{confrankone}{1st}, \colorbox{confranktwo}{2nd}, and \colorbox{confrankthree}{3rd}, where a confidence-faithful LLM should rank 1st at Top-5\% and 2nd at Top-10\%. GR@$\kappa$ is not shaded since it naturally decreases as $\kappa$ shrinks for every strategy. The best result across strategies in each column is in bold.}
\vspace{-0.3cm}
\label{app-tab:faithfulness_full}
\setlength{\tabcolsep}{5pt}
\resizebox{\linewidth}{!}{
\begin{tabular}{c l ccc|ccc}
\toprule
 & \multirow{2}{*}{\textbf{Models}} & \multicolumn{3}{c|}{\textbf{Shifting Attention to Relevance}} & \multicolumn{3}{c}{\textbf{Self-certainty}} \\
\cmidrule(lr){3-5} \cmidrule(lr){6-8}
 & & Top-5\% & Top-10\% & Top-20\% & Top-5\% & Top-10\% & Top-20\% \\
\midrule
\multirow{7}{*}{\shortstack{SelAcc@$\kappa$\\($\uparrow$)}}
& Qwen2.5-3B & \crthree{36.00} & \crtwo{56.00} & \crone{59.00} & \crtwo{56.00} & \crthree{54.00} & \crone{57.00} \\
& RD~\citep{RD} & \crone{68.00} & \crthree{62.00} & \crtwo{66.00} & \crthree{64.00} & \crtwo{66.00} & \crone{67.00} \\
& INTUITOR~\citep{zhao2025learningreasonexternalrewards} & \crthree{72.00} & \crone{80.00} & \crtwo{75.00} & \crone{\textbf{88.00}} & \crone{\textbf{88.00}} & \crthree{83.00} \\
& GRPO~\citep{shao2024deepseekmathpushinglimitsmathematical} & \crthree{56.00} & \crtwo{58.00} & \crone{65.00} & \crone{64.00} & \crthree{58.00} & \crtwo{63.00} \\
& OPSD~\cite{OPSD} & \crthree{28.00} & \crone{32.00} & \crone{32.00} & \crone{48.00} & \crthree{44.00} & \crtwo{46.00} \\
& HyTuning (Ours) & \crone{\textbf{84.00}} & \crtwo{\textbf{82.00}} & \crthree{\textbf{78.00}} & \crone{\textbf{88.00}} & \crone{\textbf{88.00}} & \crthree{\textbf{84.00}} \\
\midrule
\multirow{7}{*}{\shortstack{hAURC\\($\downarrow$)}}
& Qwen2.5-3B & \crthree{65.04} & \crtwo{60.41} & \crone{51.20} & \crone{45.30} & \crthree{47.04} & \crtwo{45.94} \\
& RD~\citep{RD} & \crone{23.76} & \crtwo{28.30} & \crthree{31.32} & \crone{28.62} & \crtwo{31.98} & \crthree{33.83} \\
& INTUITOR~\citep{zhao2025learningreasonexternalrewards} & \crthree{28.30} & \crtwo{25.42} & \crone{24.87} & \crthree{24.94} & \crtwo{19.63} & \crone{18.22} \\
& GRPO~\citep{shao2024deepseekmathpushinglimitsmathematical} & \crone{36.67} & \crthree{39.99} & \crtwo{38.27} & \crone{20.35} & \crtwo{30.72} & \crthree{34.47} \\
& OPSD~\cite{OPSD} & \crthree{74.69} & \crtwo{70.19} & \crone{68.89} & \crthree{59.70} & \crtwo{58.27} & \crone{57.37} \\
& HyTuning (Ours) & \crone{\textbf{6.77}} & \crtwo{\textbf{11.80}} & \crthree{\textbf{17.20}} & \crtwo{\textbf{12.40}} & \crone{\textbf{12.17}} & \crthree{\textbf{12.53}} \\ \midrule
\multirow{7}{*}{\shortstack{GR@$\kappa$\\($\downarrow$)}}
& Qwen2.5-3B & 3.20 & 4.40 & 8.20 & 2.20 & 4.60 & 8.60 \\
& RD~\citep{RD} & 1.60 & 3.80 & 6.80 & 1.80 & 3.40 & 6.60 \\
& INTUITOR~\citep{zhao2025learningreasonexternalrewards} & 1.40 & 2.00 & 5.00 & \textbf{0.60} & \textbf{1.20} & 3.40 \\
& GRPO~\citep{shao2024deepseekmathpushinglimitsmathematical} & 2.20 & 4.20 & 7.00 & 1.80 & 4.20 & 7.40 \\
& OPSD~\cite{OPSD} & 3.60 & 6.80 & 13.60 & 2.60 & 5.60 & 10.80 \\
& HyTuning (Ours) & \textbf{0.80} & \textbf{1.80} & \textbf{4.40} & \textbf{0.60} & \textbf{1.20} & \textbf{3.20} \\
\bottomrule
\end{tabular}}
\vspace{-0.5cm}
\end{table}

\subsection{Comprehensive Confidence Faithfulness Analysis across High-Confidence Regions}
\label{appendix:faithfulness_region}
Section~\ref{sec:conf_faith} reports hAURC within the most confident $\kappa\in\{5\%,10\%,20\%\}$ fraction of CyberMetric, and we further provide SelAcc@$\kappa$ and GR@$\kappa$ in Table~\ref{app-tab:faithfulness_full}. According to the criterion in Section~\ref{sec:Preliminary}, a confidence-faithful LLM should become increasingly reliable as $\kappa$ shrinks. Following Section~\ref{sec:conf_faith}, we regard SAR as the more objective measure, since it estimates uncertainty at the semantic level and is not directly optimized by any compared strategy, and SC as a complementary reference, since it is the intrinsic reward of both INTUITOR and the RLIF branch of HyTuning (\cf Appendix~\ref{appendix:confidence_estimation}). From Table~\ref{app-tab:faithfulness_full}, we can observe that:
(1) Under SAR, HyTuning is the only post-training strategy whose SelAcc@$\kappa$ increases monotonically as $\kappa$ shrinks, rising steadily from 78.00 at $\kappa$=20\% to 82.00 at 10\% and 84.00 at 5\%. In contrast, the SelAcc of INTUITOR fluctuates and drops to 72.00 at $\kappa\!=\!5\%$, falling below its 75.00 at 20\%, while the backbone, GRPO, and OPSD even suffer from \emph{confidence inversion}, a severe violation of confidence faithfulness where their most confident inferences are less accurate than those in the broader region. Under SC, HyTuning also exhibits a non-decreasing SelAcc from 84.00 to 88.00 and matches or exceeds INTUITOR at every $\kappa$, although the high SelAcc of INTUITOR under SC is reasonable since SC is exactly its training reward.
(2) Extending the analysis in Section~\ref{sec:conf_faith}, the confidence inversion is also reflected by hAURC, which rises for INTUITOR, the backbone, and OPSD as $\kappa$ shrinks from 20\% to 5\%. RD is the only baseline whose hAURC decreases monotonically like HyTuning, yet its value at $\kappa\!=\!$5\% remains more than three times that of HyTuning (23.76 \emph{vs.} 6.77). A similar contrast appears under SC, where the hAURC of INTUITOR rises from 18.22 to 24.94 even though SC is its own training reward, while that of HyTuning stays the lowest at around 12\%.
(3) Since GR@$\kappa$ is normalized by all the queries, it naturally decreases as $\kappa$ shrinks for every post-training strategy, while directly reflecting how often a deployed LLM delivers confidently-wrong inferences to practitioners. Under this metric, HyTuning attains the lowest GR at every $\kappa$ under both confidence measures, matched only by INTUITOR under SC at $\kappa\!=\!5\%$.
(4) The contrast between SAR and SC further distinguishes the two post-training strategies that optimize SC. The top-5\% SelAcc of INTUITOR drops from 88.00 under SC to 72.00 under SAR, whereas that of HyTuning only slightly decreases from 88.00 to 84.00. This indicates that the confidence optimized by INTUITOR fails to transfer to the independent semantic-level measure, while that of HyTuning remains consistent across both.
Overall, the advantage of HyTuning holds consistently across the examined high-confidence regions under both measures, and its confidence becomes increasingly trustworthy exactly where high-stakes deployment relies on it most.

\FloatBarrier
\begin{table*}[!t]
\caption{Case study of HyTuning and INTUITOR on CyberMetric.}
\vspace{-0.2cm}
\label{tab:case_study_cybermetric}
\normalsize
\centering
\resizebox{\linewidth}{!}{
\begin{tabular}{@{}c|c@{}}
\toprule
\multicolumn{2}{@{}c@{}}{
  \parbox{\linewidth}{
    \textbf{Question}: What is the primary purpose of the traceroute program in network testing?
    \\[0.3em]
\begin{tabular}[t]{@{}l@{\quad}p{0.8\linewidth}@{}}
  \textbf{Answer Choices}:
  & A. To monitor and control internet traffic flow\\
  & B. To identify the physical location of network routers\\
  & C. To encrypt and secure data transmissions within the network\\
  & \textcolor{ForestGreen}{
      \textbf{D. To determine the path and time taken for
      data packets to reach their destination}
      (\textbf{Correct Answer})
    }
\end{tabular}
  }
} \\
\midrule
\textbf{HyTuning} & \textbf{INTUITOR} \\
\midrule
\parbox[t]{0.46\linewidth}{
To determine the primary purpose of the traceroute program, let's
analyze each option and its relevance to network testing:
\par\medskip

A. To monitor and control internet traffic flow. This option is related to network management and monitoring, but it does not specifically address the purpose of traceroute. \textcolor{ForestGreen}{Traceroute is \textbf{more focused on} identifying the path and time taken for data packets to reach their destination.}
\par\medskip

B. To identify the physical location of network routers. \textcolor{ForestGreen}{Traceroute is \textbf{primarily used to} determine the path and time taken for data packets to reach their destination. It does not directly provide information about the physical location of network routers.}
\par\medskip

C. To encrypt and secure data transmissions within the network. This option is related to network security, but it does not specifically address the purpose of traceroute. \textcolor{ForestGreen}{Traceroute is not used for encrypting or securing data transmissions.}
\par\medskip

D. To determine the path and time taken for data packets to reach their destination. \textcolor{ForestGreen}{\textbf{This option accurately describes the primary purpose of the traceroute program.} Traceroute is \textbf{a network diagnostic tool that provides information} about the path and time taken for data packets to reach their destination.} It helps identify any potential issues or bottlenecks in the network.
\par\medskip

Based on this analysis, the best answer is:
\par\medskip

\textcolor{ForestGreen}{Answer : D \textbf{(Correct Answer)}}
}
&
\parbox[t]{0.54\linewidth}{
To determine the primary purpose of the traceroute program in network testing, let's analyze each option step by step:
\par\medskip

A. To monitor and control internet traffic flow. This option is incorrect because the traceroute program is not designed to monitor or control internet traffic flow. Its primary purpose is to trace the path that data packets take to reach their destination.
\par\medskip

B. To identify the physical location of network routers. \textcolor{red}{\textbf{This option is correct.} The traceroute program is specifically designed to \textbf{identify the physical location of network routers} along the path that data packets take to reach their destination.} It does this by sending data packets to a destination and recording the responses from each router along the path.
\par\medskip

C. To encrypt and secure data transmissions within the network. This option is incorrect because the traceroute program is not designed to encrypt or secure data transmissions. Its primary purpose is to trace the path that data packets take to reach their destination.
\par\medskip

D. To determine the path and time taken for data packets to reach their destination. \textcolor{red}{\textbf{This option is correct.} The traceroute program is designed to \textbf{determine the path and time taken} for data packets to reach their destination.} It does this by sending data packets to a destination and recording the responses from each router along the path.
\par\medskip

Based on the analysis, \textcolor{red}{the primary purpose of the traceroute program in network testing is to \textbf{identify the physical location of network routers} along the path that data packets take to reach their destination.}
\par\medskip
\textcolor{red}{Answer: B \textbf{(Incorrect Answer)}}
} \\
\bottomrule
\end{tabular}
}
\vspace{-0.5cm}
\end{table*}

\FloatBarrier
\begin{table*}[!t]
\caption{Case study of HyTuning and INTUITOR on MMLU.}
\vspace{-0.2cm}
\label{tab:case_study_mmlu}
\normalsize
\centering
\resizebox{\linewidth}{!}{
\begin{tabular}{@{}c|c@{}}
\toprule
\multicolumn{2}{@{}c@{}}{
  \parbox{\linewidth}{
    \textbf{Question}: Given that the sample has a standard deviation
    of zero, which of the following is a true statement?
    \\[0.3em]
    \begin{tabular}[t]{@{}l@{\quad}p{0.8\linewidth}@{}}
      \textbf{Answer Choices}:
      & A. The standard deviation of the population is also zero.\\
      & \textcolor{ForestGreen}{
          \textbf{B. The sample mean and sample median are equal.}
          (\textbf{Correct Answer})
        }\\
      & C. The sample may have outliers.\\
      & D. The population has a symmetric distribution.
    \end{tabular}
  }
} \\
\midrule
\textbf{HyTuning} & \textbf{INTUITOR} \\
\midrule
\parbox[t]{0.48\linewidth}{
Let's analyze the given question and the provided options step by
step to determine the correct answer.
\par\medskip

1. Understanding Standard Deviation:
The standard deviation measures the amount of variation or
dispersion in a set of values.
\textcolor{ForestGreen}{A standard deviation of zero indicates
that \textbf{all the values in the sample are identical}.}
\par\medskip

2. Implications of Zero Standard Deviation:
If the standard deviation of the sample is zero, it means that
every data point in the sample is the same. This implies that
the sample is perfectly homogeneous.
\par\medskip

A. The standard deviation of the population is also zero.
\textcolor{ForestGreen}{\textbf{This statement is not necessarily
true.} The standard deviation of the population depends on the
variability of the entire population, not just the sample.
\textbf{Even if the sample has a standard deviation of zero,
the population could still have variability.}}
\par\medskip

B. The sample mean and sample median are equal.
\textcolor{ForestGreen}{In a perfectly homogeneous sample,
\textbf{the mean and median are indeed equal}. This is because
there is only one unique value, and both measures of central
tendency will be that value.}
\par\medskip

C. The sample may have outliers.
This statement is false. If the sample has a standard deviation
of zero, it means there are no differences among the values,
so there cannot be any outliers.
\par\medskip

D. The population has a symmetric distribution.
This statement is not necessarily true. A symmetric distribution
is characterized by equal probabilities on both sides of the
mean. However, a perfectly homogeneous sample does not imply
a symmetric distribution of the population. The population
could have a variety of distributions, including skewed or
uniform distributions.
\par\medskip

Based on the analysis, the only true statement is that the
sample mean and sample median are equal.
\par\medskip

\textcolor{ForestGreen}{Answer : B \textbf{(Correct Answer)}}
}
&
\parbox[t]{0.52\linewidth}{
To determine which of the given statements is true given that
the sample has a standard deviation of zero, let's analyze
each option step by step.
\par\medskip

Given: The sample has a standard deviation of zero.
\par\medskip

A. The standard deviation of the population is also zero.
The standard deviation of a sample is an estimate of the
standard deviation of the population. If the sample standard
deviation is zero, it suggests that all the data points in
the sample are identical.
\textcolor{red}{\textbf{This implies that the population standard
deviation is also zero. Therefore, this statement is true.}}
\par\medskip

B. The sample mean and sample median are equal.
If the sample standard deviation is zero, it means that all
the data points in the sample are identical.
\textbf{In such a case, the sample mean and the sample median
are both equal to the value of the data points. Therefore,
this statement is true.}
\par\medskip

C. The sample may have outliers.
If the sample standard deviation is zero, it means that all
the data points in the sample are identical. Outliers are
data points that are significantly different from the rest
of the data. Since all data points are identical, there are
no outliers. Therefore, this statement is false.
\par\medskip

D. The population has a symmetric distribution.
If the sample standard deviation is zero, it means that all
the data points in the sample are identical. This does not
necessarily imply that the population has a symmetric
distribution. The population could have a symmetric
distribution, but it could also have a distribution with
a single value repeated. Therefore, this statement is
not necessarily true.
\par\medskip

Based on the analysis, the true statements are:
\par
Option A: The standard deviation of the population is also zero.
\par
Option B: The sample mean and sample median are equal.
\par\medskip

\textcolor{red}{However, since we need to choose the best answer, and \textbf{both A and B are true, we can select either one. For the sake of this response, I will choose the first true statement}:}
\par\medskip

\textcolor{red}{Answer: A \textbf{(Incorrect Answer)}}
} \\
\bottomrule
\end{tabular}
}
\vspace{-0.5cm}
\end{table*}

\subsection{Case Studies}
\label{appendix:case_study}
To complement the quantitative analysis of confidence faithfulness in Section~\ref{sec:conf_faith} and Appendix~\ref{appendix:faithfulness_region}, we present two cases that compare HyTuning with INTUITOR on the same queries. We mainly focus on a CyberMetric case from the high-stakes domain (\cf Table~\ref{tab:case_study_cybermetric}), and further include an MMLU case from the general domain (\cf Table~\ref{tab:case_study_mmlu}) to show that the observed pattern is not confined to domain knowledge. 
Rather than contrasting the correctness of final answers, we examine how each model reaches its commitment along the reasoning trace. For readability, we omit the text generated after the final answer, the verbatim restatement of the query, and the Markdown formatting.

\textbf{Case in the High-Stakes Domain.} Table~\ref{tab:case_study_cybermetric} presents a question in CyberMetric on the primary purpose of traceroute program. HyTuning expresses increasing confidence in the function of traceroute as its reasoning unfolds. When analyzing the distractor A, it mentions this function only with the tentative wording ``\textbf{is more focused on}''. When analyzing option B, the wording strengthens to ``\textbf{is primarily used to}'', and HyTuning explicitly rules out the physical location of routers. When analyzing option D, it asserts that traceroute ``\textbf{is a network diagnostic tool}'' and commits to the answer only after this step. \textbf{This progression from tentative to definitive wording is consistent with the premise of PRG that sound reasoning moves from uncertainty to certainty} (\cf Section~\ref{sec:Preliminary}). In contrast, INTUITOR exhibits factually unwarranted overconfidence that is decoupled from its own reasoning. It declares ``This option is correct'' for both B and D and its analyses of A and C further state twice that the primary purpose of traceroute is to trace the path of data packets, which directly support D. Nevertheless, INTUITOR commits to B and folds a fabricated physical-location function into its conclusion. \textbf{Its certainty in B thus stems from a single assertion rather than accumulating step by step from the evidence it generates, which is precisely the pattern that PRG is designed to identify.} In a high-stakes deployment, such a definitive yet self-contradictory response may mislead practitioners. This contrast aligns with the PRG-based weighting of HyTuning, which dynamically attenuates intrinsic sharpening on trajectories without accumulated confidence and shifts more weight toward the verified target whenever one is available.

\textbf{Case in the General Domain.} Table~\ref{tab:case_study_mmlu} presents a question in MMLU on a sample with zero standard deviation. Both models derive the same premise that all the values in the sample are identical and they diverge only on whether this premise extends from the sample to the population. INTUITOR extends it to the population and asserts that option A is true. It then declares B true as well and breaks the tie simply by option order, outputting a definitive answer that is determined by position rather than by reasoning. This behavior is in line with the observation that the standard post-training and evaluation places a correct guess on the same footing as a reasoned answer~\citep{kalai2025language}. In contrast, HyTuning explicitly states that the population could still have variability. It grades each option as ``\textbf{not necessarily true}'', ``\textbf{false}'', or ``\textbf{indeed equal}'' according to its logical status, and reaches ``\textbf{the only true statement}'' only after ruling out the alternatives. Together with the MMLU accuracy in Table~\ref{tab.overall_performance}, this case suggests that HyTuning keeps its commitments within the support of its reasoning beyond the high-stakes domain.
\clearpage
\section{Prompt Templates}
\label{app-sec:prompt}

\begin{figure*}[!t]
\footnotesize
\refstepcounter{myboxcounter}\label{prompt:generate_syn_data}
\begin{custombox_black}[top=1mm, bottom=1mm, left=1mm, right=1mm]{Prompt template for \emph{synthetic data generation}. }
\textbf{[System]:} You are an expert cybersecurity data generator and analyst.\\
\textbf{[User]:} Your task is to synthesize new cybersecurity data samples that are realistic, diverse, and consistent with the structure and semantics of the provided few-shot examples.
\\\\
The new samples must represent the SAME TYPES OF TASKS as the examples:\\
- cwe\_map: CWE (Common Weakness Enumeration) mapping\\
- calc\_cvss: CVSS (Common Vulnerability Scoring System) calculation or estimation\\
- mcq: Multiple-choice questions (MCQ) testing cybersecurity or CTI (Cyber Threat Intelligence) knowledge\\
- mitre\_map: MITRE ATT\&CK technique mapping\\
\\
IMPORTANT:\\
You must ONLY generate the \textbf{input/question part} of each data sample. DO NOT include or reveal the answer, label, output, or solution.
For example:\\
- For CWE mapping, provide only the vulnerability description, not the mapped CWE ID.\\
- For CVSS calculation, describe the vulnerability scenario, not the CVSS score.\\
- For MCQs, provide only the question text and answer choices, not the correct answer.\\
- For MITRE mapping, describe the adversary behavior, not the MITRE ID or tactic.\\
\\
Follow these rules carefully:\\
1. Maintain the same overall structure, schema, and output format as the few-shot examples.\\
2. Keep field names identical and values consistent with the type of data shown in the examples.\\
3. Ensure all generated data is realistic and domain-accurate:\\
- Use valid CVSS scores and components.\\
- Use real-looking CWEs (e.g. CWE-79, CWE-89) and MITRE techniques (e.g. T1059, T1204).\\
- Use plausible cybersecurity contexts, scenarios, or vulnerabilities.\\
4. Do NOT copy or rephrase the few-shot examples. Create entirely new but believable samples.\\
5. Do NOT include the answer, label, output or solution for each data sample.\\
6. Maintain internal consistency:\\
- If the CWE corresponds to a certain vulnerability type, ensure the description aligns with it.\\
- CVSS metrics should make sense.\\
7. Ensure task diversity:\\
- Include multiple task types (CWE mapping, CVSS scoring, MCQs, MITRE mappings) in the new samples.\\
- Vary topics, severities, and threat types.\\
8. Output formatting:\\
- Each generated data sample MUST have their task\_type clearly stated at the beginning of the sample:\\
task\_type: (cwe\_map OR calc\_cvss OR mcq OR mitre\_map)\\
- Each generated data sample MUST be clearly separated by delimiter lines:\\
===== SAMPLE BEGIN =====\\
===== SAMPLE END =====\\

Few-shot examples: \{\uline{Few\_Shot\_Examples}\}\\
        
Task: Generate 10 new, synthetic cybersecurity data samples following the same schema, tone and task categories.
\end{custombox_black}
\end{figure*}

\subsection{Prompt Template for Synthetic Data Generation}
\label{app:Prompt_Data_Generation}
Under partial supervision, we construct the unlabeled questions for the RLIF branch of HyTuning by prompting \texttt{Qwen2.5-32B-Instruct} with the template in Box~\ref{prompt:generate_syn_data}, where the randomly-sampled 100 supervised queries serve as few-shot seeds (\cf Appendix~\ref{appendix:training_corpus}). The template requires the LLM to generate only questions without their corresponding answers, which keeps the synthetic corpus label-free as the RLIF branch of HyTuning requires. Besides, it constrains the generated questions to share the task types and output schema of the seed questions while remaining realistic, diverse, and distinct from them.

\subsection{Prompt Template for Quality Assessment}
\label{app:Prompt_Sample_Quality}
Box~\ref{prompt:data_filter_quality_eval} presents the judging prompt that instantiates the criteria specified by three cybersecurity experts to select the Top-100 supervised traces in Appendix~\ref{app:top100_training_corpus}. Specifically, it rates each instance on relevance, answer quality, and language quality, and then assigns an overall label. We adopt \texttt{GPT-5} as the judge and retain only the samples rated high on all three aspects as candidates.

\begin{figure*}[!t]
\centering
\footnotesize
\refstepcounter{myboxcounter}\label{prompt:data_filter_quality_eval}
\begin{custombox_black}[top=1mm, bottom=1mm, left=1mm, right=1mm]{Prompt template for \emph{evaluating the quality of the Primus-Reasoning dataset}. }
\textbf{[System]:} You are an expert cybersecurity data analyst.\\
\textbf{[User]:} You will be given a question and an answer related to cybersecurity topics such as:\\
- CWE (Common Weakness Enumeration) mapping\\
- CVSS (Common Vulnerability Scoring System) score calculation\\
- Multiple-choice questions (MCQ) testing cybersecurity or CTI (Cyber Threat Intelligence) knowledge\\
- MITRE ATT\&CK technique mapping\\

Your task is to assess the \textbf{quality} of the provided Q\&A sample. For each input, perform the following steps:

1. \textbf{Relevance Check}:\\
- Is the question relevant to cybersecurity, vulnerability analysis, or CTI?
- Does it relate logically to topics like CWE, CVSS, MITRE, etc.?\\
2. \textbf{Answer Quality}:\\
- Is the answer factually correct, coherent, and complete?\\
- Does it show technical accuracy and clear reasoning?
\\
3. \textbf{Language and Structure}:\\
- Is the question clearly phrased and understandable?\\
- Is the answer grammatically correct and concise?
\\
4. \textbf{Overall Label}:\\
- Label the sample as one of:
    - 'high\_quality' (technically accurate, relevant, clear)\\
    - 'medium\_quality' (some minor issues but usable)\\
    - 'low\_quality' (irrelevant, incoherent, incorrect, or unusable)
\\\\
Return your output as a JSON object with the following structure:
\{\{
"relevance": "< High | Medium | Low >",
"answer\_quality": "< High | Medium | Low >",
"language\_quality": "< High | Medium | Low >",
"overall\_label": "<high\_quality | medium\_quality | low\_quality>",
\}\}

Question:
\{\uline{Question}\}

Answer:
\{\uline{Answer}\}
\end{custombox_black}
\end{figure*}

\end{document}